\pdfoutput=1

\documentclass[11pt]{article}

\usepackage{acl}

\usepackage{times}
\usepackage[utf8]{inputenc}
\usepackage{latexsym}
\usepackage{graphicx}
\usepackage{colortbl}
\usepackage{multirow}
\usepackage{booktabs}
\usepackage{subcaption}
\usepackage{siunitx}
\usepackage[export]{adjustbox}
\usepackage{dirtytalk}
\usepackage{booktabs}
\usepackage{array}
\usepackage{xcolor}
\usepackage{colortbl}
\usepackage{makecell}
\usepackage{adjustbox}
\usepackage{ragged2e}
\usepackage{xcolor}
\usepackage{colortbl}
\usepackage{array}
\usepackage{graphicx}
\usepackage{amsmath}

\definecolor{repetitive}{RGB}{255, 240, 200}    
\definecolor{erratic}{RGB}{255, 220, 220}       
\definecolor{repetitive}{RGB}{230, 225, 240}    
\definecolor{erratic}{RGB}{245, 225, 215}       

\definecolor{headercolor}{RGB}{70, 130, 180}
\definecolor{rowcolor}{RGB}{240, 248, 255}
\usepackage{algorithm}
\usepackage{algpseudocode}
\usepackage{float}
\usepackage{pdflscape}
\usepackage{hyperref}
\usepackage[]{acronym}
\usepackage{amsmath}
\usepackage{amssymb}
\usepackage{amsfonts}

\usepackage{amssymb}

\usepackage{comment}
\usepackage[T1]{fontenc}

\usepackage[utf8]{inputenc}

\usepackage{microtype}

\usepackage{inconsolata}

\usepackage{graphicx}

\title{Towards Better Open-Ended Text Generation: \\ A Multicriteria Evaluation Framework}

\author{
  \textbf{Esteban Garces Arias\textsuperscript{1,2}},
    \textbf{Hannah Blocher\textsuperscript{1}},
  \textbf{Julian Rodemann\textsuperscript{1}},
  \textbf{Meimingwei Li\textsuperscript{1}},\\
  \textbf{Christian Heumann\textsuperscript{1}},
  \textbf{Matthias Aßenmacher\textsuperscript{1,2}}
\\
\\
  \textsuperscript{1}Department of Statistics, LMU Munich,\\
  \textsuperscript{2}Munich Center for Machine Learning (MCML)
\\
\\
  \small{
    \textbf{Correspondence:} \href{mailto:Esteban.GarcesArias@stat.uni-muenchen.de}{Esteban.GarcesArias@stat.uni-muenchen.de}
  }
}

\begin{document}

\maketitle

\begin{abstract} 
Open-ended text generation has become a prominent task in natural language processing due to the rise of powerful (large) language models. However, evaluating the quality of these models and the employed decoding strategies remains challenging due to trade-offs among widely used metrics such as coherence, diversity, and perplexity. This paper addresses the specific problem of multicriteria evaluation for open-ended text generation, proposing novel methods for both relative and absolute rankings of decoding methods. Specifically, we employ benchmarking approaches based on partial orderings and present a new summary metric to balance existing automatic indicators, providing a more holistic evaluation of text generation quality. Our experiments demonstrate that the proposed approaches offer a robust way to compare decoding strategies and serve as valuable tools to guide model selection for open-ended text generation tasks. We suggest future directions for improving evaluation methodologies in text generation and make our code, datasets, and models publicly available.\footnote{\url{https://github.com/YecanLee/2BeOETG}}
\end{abstract}

\section{Introduction}\label{sec:intro}

Large language models \citep[LLMs, e.g.,][]{dubey2024llama3herdmodels,yang2024qwen2technicalreport} have demonstrated remarkable capabilities in generating coherent and contextually appropriate text across diverse domains. However, the quality of LLM outputs is fundamentally determined not only by the underlying model architecture but also by the decoding strategies employed during inference—the algorithms that transform the model's output probability distributions into actual text sequences. As the landscape of both LLMs and decoding strategies continues to expand rapidly, the need for robust evaluation frameworks has become increasingly critical \citep{wiher-etal-2022-decoding, garces-arias-etal-2025-decoding}.

\begin{figure}[ht]
    \centering 
    \includegraphics[width=80mm, keepaspectratio]{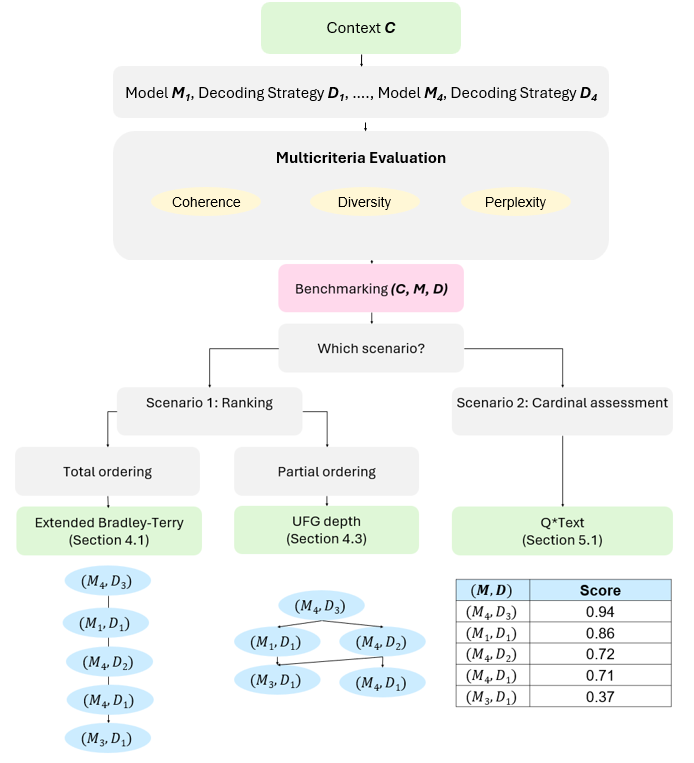} 
    \caption{\textbf{Multicriteria evaluation framework} for benchmarking models and decoding strategies, i.e., \textit{decoding methods}. We distinguish two scenarios for benchmarking (\S\ref{sec:intro}) and two ranking objectives (\S\ref{sec:benchmarking}), giving rise to three use-case tailored, distinct methods (\S\ref{sec:bradley_terry},~\ref{sec:ufg}~and~\ref{sec:collapsing_single_metric}).} 
    \label{fig:introduction} 
\end{figure}

\paragraph{Scope and Problem Definition.} This paper specifically addresses the challenge of multicriteria evaluation in open-ended text generation, where we must simultaneously consider multiple, often conflicting quality dimensions \citep{holtzman2019curious, su2022empirical}. We focus on developing principled methods for both relative and absolute rankings of decoding methods. Our approach centers on a subset of automatic evaluation metrics—coherence, diversity, and generation perplexity—that capture fundamental trade-offs in text generation quality. While numerous other metrics exist (e.g., relevance, informativeness, style consistency), we deliberately limit our scope to these three core dimensions to establish a foundational framework that can be systematically extended.

Current evaluation approaches face remarkable limitations when assessing the quality of text generations within this multicriteria context. Traditional methods typically rely on either human judgments—considered the gold standard, but resource-intensive, and dependent on carefully designed protocols \citep{howcroft-etal-2020-twenty, VANDERLEE2021101151, karpinska-etal-2021-perils, ruan-etal-2024-defining}—or individual automatic metrics. While automatic metrics such as MAUVE~\citep{pillutla2021mauve}, coherence~\citep{su2022contrastive}, diversity, and generation perplexity~\citep{10.1121/1.2016299} provide valuable insights into specific aspects of generation quality, an isolated consideration of these measures offers only an incomplete perspective on overall performance and fails to address the fundamental multicriteria nature of the evaluation problem.

In the context of open-ended text generation, this evaluation challenge is particularly acute because decoding strategies inherently involve trade-offs between competing objectives such as coherence and diversity. A method that excels in coherence may underperform in diversity, and vice versa, making it difficult to establish consistent relative rankings among different approaches or provide meaningful absolute assessments of their quality.

The fundamental challenge addressed in this work lies in developing principled approaches for both relative and absolute multicriteria evaluation that can balance our selected subset of automatic metrics within a comprehensive framework. This enables reliable comparison of different models and decoding strategies—collectively referred to as \textit{decoding methods} throughout this work (Fig.~\ref{fig:introduction})—while acknowledging the inherent trade-offs between the chosen evaluation criteria. Addressing this challenge is essential for advancing the field of open-ended text generation evaluation and providing practitioners with evidence-based guidance for selecting optimal decoding methods within the multicriteria landscape we define.

\paragraph{Research Gap.} When evaluating decoding methods based on multiple quality criteria in several scenarios (i.e., datasets), a method may excel in one area while lagging in another. Aggregating such \textit{multicriteria evaluation results} for different scenarios is still an open problem. Existing approaches comprise the Pareto front or weighted sums. While the former is hardly informative for large-scale benchmarking (cf. \S\ref{sec:benchmarking}), the latter depends on (arbitrarily) selected weights. In this work, we offer two alternative approaches while distinguishing two\footnote{In reality, one can imagine a multitude of scenarios in between these two prototypical cases, hence we also consider benchmarking methods along this spectrum. What unites them, however, is their ability to aggregate multiple criteria.} prototypical \textit{practical} benchmarking scenarios with associated \textbf{research questions (RQ)}: 

\paragraph{Scenario 1 (Ranking).} First, consider a practitioner using open-ended text generation for a specific task, e.g., a customer support chatbot. This practitioner might primarily be interested in a complete scenario-specific relative ranking of existing methods. This motivation renders metric information about the methods' performances a means to an end. Thus, an \textit{ordinal ranking} of methods will do. \textbf{RQ1:} Can we exploit novel statistical methodologies for partial orders to establish \textit{multicriteria} rankings that potentially allow for incomparability?

\paragraph{Scenario 2 (Cardinal Assessment).} Second, for researchers interested in designing new decoding methods (i.e., model, decoding strategy, or both), it is of utmost importance to know \textit{how much} better one method is compared to another, i.e., having \textit{an absolute ranking on a cardinal scale}. Knowledge of the performance of existing methods on different tasks will help derive new methods. \textbf{RQ2:} Can we aggregate multiple automatic evaluation metrics in a meaningful and statistically valid way? 

\paragraph{Contributions.} We address \textbf{RQ1} (\S\ref{sec:benchmarking}) and \textbf{RQ2} (\S\ref{sec:collapsing_single_metric}) by proposing appropriate aggregation methods (cf. Fig.~\ref{fig:introduction}), including a novel summary metric to balance multiple assessments. We further provide experimental results by applying all introduced methods to over 1.8M stories generated by six LLMs on real-world datasets (cf. \S\ref{sec:experimental_setup} for the setup and \S\ref{sec:ebt_results}, \S\ref{sec:ufg_results}, \S\ref{sec:qtext_results} for the results).

\section{Related Work}
\label{sec:related_work}

Benchmarks are ubiquitous in applied machine learning (ML) research \cite{zhang2024inherent,shirali2023theory,ott2022mapping,zhang2020machine,thiyagalingam2022scientific,roelofs2019meta,vanschoren2014openml}, being used to make informed decisions and to demonstrate the superiority of newly proposed methods over concurrent ones \cite{MEYER2003169,Hothorn,ehl2012,mptbw2015}. In recent years, the focus has shifted towards multicriteria and multi-task benchmarking problems \cite{cruz2024evaluating, zhanginherent,kohli2024towards,jansen2024statistical,jansen2023statistical,jansen2023robust,rodemann2024partial,blocher2024comparing}. In a multitude of domains, there are several criteria concerning which methods need to be compared. Classical examples include runtime and accuracy in predictive ML \cite{koch2015efficient, jansen2024statistical} or performance and speed in optimization \cite{schneider2018deepobs}, to name only a few.

Modern LLMs require evaluation across multiple metrics due to their broad capabilities \citep[see, e.g.,][]{wei2024rethinking,liu2025reevaluating}. Assessing models on diverse tasks -- from reasoning and comprehension to creativity and ethics -- provides better understanding of their strengths and limitations \cite{chiang2024chatbot}. These comprehensive evaluation frameworks advance model performance while ensuring alignment with real-world applications and ethical standards \cite{liu2023alignbench,ji2023ai,terry2023ai,pos-25}. Multicriteria benchmarking has thus become essential for guiding both theoretical progress and practical deployment of LLMs.

\begin{table*}[!ht]
\centering
\resizebox{1\textwidth}{!}{
\begin{tabular}{llllllr}
\hline
\textbf{Models} & \textbf{Datasets} & \textbf{Metrics} & \textbf{Decoding strategy}  & \textbf{Hyperparameter} & \textbf{Values}                  & \multicolumn{1}{l}{\textbf{\# Data points}} \\ \hline\hline
Deepseek        & Wikitext          & Coherence        & Beam search                 & $B$              & \{3, 5, 10, 15, 20, 50\}      & 6 × 5261 × 6 = 189,396                             \\
Falcon2          & Wikinews          & Diversity        & Contrastive search          & $k$                       & \{1, 3, 5, 10, 15, 20, 50\}      & 6 × 5261 × 7 × 5 = 1,104,810                       \\
GPT2-XL         & Book              & Gen. Perplexity            &                             & $\alpha$                   & \{0.2, 0.4, 0.6, 0.8, 1.0\}      &                                             \\
Llama3          &                   &  & Temperature sampling & $\tau$                       & \{0.1, 0.3, 0.5, 0.7, 0.9, 1.0\}                             & 6 × 5261 × 6 = 189,396                             \\
Mistralv03          &                   &                  & Top-$k$ sampling   & $k$              & \{1, 3, 5, 10, 15, 20, 50\}  & 6 x 5261 x 7 = 220,962                                \\
Qwen2     &                   &                  & Top-$p$ (nucleus) sampling              & $p$                        & \{0.6, 0.7, 0.8, 0.9, 0.95\}      & 6 × 5261 × 5 = 157,830                             \\
  \hline
                &                   &                  &                             &                         & Grand Total                      & \multicolumn{1}{r}{1,862,394}                  
\end{tabular}
}
\caption{Experimental setup: Over 1.8M text generations produced using various models and decoding strategies with different hyperparameter configurations. Prompts were drawn from three datasets (Wikitext, Wikinews, and Book), and outputs were evaluated on Coherence, Diversity, and Generation Perplexity.}
\label{tab:experimental_setup}
\end{table*}

Decoding methods for open-ended text generation are no exception. Several metrics to evaluate the quality of decoding strategies have been proposed and discussed in recent years \cite{alihosseini-etal-2019-jointly,celikyilmaz2021evaluationtextgenerationsurvey,su2022empirical,su2022contrastive,gao2022simcse, becker2024textgenerationsystematicliterature,garces-arias-etal-2025-decoding}. Diversity, MAUVE, coherence, and generation perplexity are among the most popular metrics. Diversity measures lexical variation using $n$-gram repetition rates, with higher scores indicating less repetition. MAUVE is a distribution similarity metric between generations and reference texts. Coherence is defined as the averaged log-likelihood of the generated text conditioned on the prompt and rewards logical flow. Finally, generation perplexity \cite{10.1121/1.2016299} measures the predictability of the generated text under the language model; lower perplexity indicates that the text is more likely according to the model's own probability distribution.

This multitude of quality metrics naturally raises the question of how to aggregate them, i.e., how to account for multiple dimensions of text quality to compare decoding methods holistically. It is self-evident that focusing on single metrics has obvious shortcomings. Exclusively optimizing for coherence will favor decoding methods with only moderate diversity, leading to \textit{degenerate}, i.e., repetitive and uncreative generations \citep{holtzman2019curious, lee2022factuality}. On the other hand, focusing solely on diversity will eventually result in incoherent text only slightly -- if at all -- related to the prompt. In this work, we offer a fresh perspective on the problem of multicriteria evaluation, adopting recent developments in the theory of depth functions and order theory (cf. \S\ref{sec:benchmarking}).

\section{Experimental Setup}
\label{sec:experimental_setup}

We evaluate six model architectures that generated over 1.8 million stories based on prompts sourced from three distinct datasets, utilizing five decoding strategies across 59 hyperparameter configurations.

\paragraph{Models.} We employ GPT2-XL \citep[1.5B,][]{radford2019language}, Mistral 7B v0.3 \citep{jiang2023mistral7b}, Llama 3.1 8B \citep{dubey2024llama3herdmodels}, Deepseek 7B \citep{deepseekai2024deepseekllmscalingopensource}, Qwen 2 7B \citep{yang2024qwen2technicalreport}, and Falcon 2 11B \citep{malartic2024falcon211btechnicalreport}. 

\paragraph{Evaluation Metrics.} Building upon \citet{su2023contrastive}, we select diversity, coherence, and generation perplexity\footnote{For their definitions, please refer to Appendix \ref{a:autom_metrics}.} as automatic metrics to assess the quality of the generated texts individually. Based on this subset of possible instance-level metrics, we construct partial orders for multicriteria rankings (\S\ref{sec:benchmarking}) and develop a cardinal assessment that collapses all metrics into one single score (\S\ref{sec:collapsing_single_metric}). Since both approaches require instance-level metrics, we exclude MAUVE in this study as it assesses distributional similarities between samples of machine-generated text and human-written continuations, i.e. it relies on aggregated data, which would prevent us from applying the methods proposed in \S\ref{sec:benchmarking} and \S\ref{sec:collapsing_single_metric}.

\paragraph{Datasets.} We evaluate our methods across three domains for open-ended text generation: News, Wikipedia articles, and stories. Specifically, we use 2,000 articles from Wikinews for the news domain; 1,314 articles from the WikiText-103 dataset \citep{merity2016pointer} for the Wikipedia domain; and 1,947 examples from the Project Gutenberg split of the BookCorpus \citep{zhu2015aligning} for the story domain. Each example consists of a prompt and a gold reference (i.e., a human continuation) for evaluation. Further, we utilize the dataset provided by \citet{garces-arias-etal-2025-decoding}, including over 1.8M generated continuations (with a maximum length of 256 tokens) for each prompt, along with aggregated metrics (coherence, diversity, MAUVE). We extend this dataset by computing sentence-level metrics and incorporating generation perplexity.

\paragraph{Decoding Strategies and Hyperparameters.}
For contrastive search \citep[CS,][]{su2022contrastive}, we evaluate 35 combinations of $\alpha$ and $k$, while for beam search \citep[BS,][]{Freitag_2017}, we consider six beam widths $B$. For temperature sampling \citep{ackley1985learning}, we consider six different temperatures $\tau$, for top-$k$ sampling \citep{fan2018hierarchical}, we use 7 different $k$ values and for top-$p$ (nucleus) sampling \citep{holtzman2019curious} we evaluate five different values for $p$, for a total of 59 decoding strategies configurations. All details are summarized in Table \ref{tab:experimental_setup}.

\section{Scenario 1: Ranking Methods}\label{sec:benchmarking}

To benchmark decoding methods according to multiple criteria (cf. \S\ref{sec:related_work}) aiming for a ranking of methods (Scenario 1 and \textbf{RQ1} in \S\ref{sec:intro}), we adopt very recent developments in the theory of multicriteria and multitask benchmarking \cite{jansen2023robust,jansen2023statistical,cruz2024evaluating, zhanginherent,kohli2024towards,jansen2024statistical,rodemann2024partial,blocher2024comparing}, some of them grounded in decision theory (social choice theory), some in the theory of data depth.

In this section, we propose benchmarking of decoding methods in terms of an \textit{ordinal ranking} along (i) the extended Bradley-Terry model \citep[\S\ref{sec:bradley_terry};][]{bradley-terry-model} and (ii) the union-free-generic (ufg) depth \citep[\S\ref{sec:ufg};][]{blocher2024comparing,blocher2024data} as an alternative approach. Both approaches deliver ordinal rankings of decoding methods rather than a cardinal quality assessment (cf. left and middle column of Table \ref{tab:comparison_methods}). This can be motivated from a practical perspective (cf. \S\ref{sec:intro}): The cardinal information incorporated in numerous metrics can be considered redundant in cases when pure \textit{ranking} of the decoding methods is the overall aim of benchmarking, not assigning scores to them. After all, a decoding method can either be deployed by practitioners or not, rendering the metric information not of primary practical interest. 

\paragraph{Use Case}

To illustrate our evaluation methodology, we apply it to the WikiText-103 dataset, which comprises 1,314 human-written prompts. We assess decoding methods by analyzing their text generations across three quality metrics: coherence, generation perplexity, and diversity. Our benchmarking approach produces partial rankings by determining whether one decoding method outperforms another, without quantifying the magnitude of performance differences.

Given the use of multiple quality metrics, we employ a dominance-based comparison framework. A decoding method is considered superior to another if and only if all three metrics either support this preference or remain neutral (i.e., do not contradict it). Consider, for example, the performance of Mistral 3 CS with hyperparameter configurations (('0.2', '1')) and (('0.8', '1')) on the first WikiText prompt. We observe that the coherence metric demonstrates a strict preference for (('0.2', '1')) over (('0.8', '1')), while the perplexity and diversity metrics show no contradictory evidence. Consequently, we conclude that Mistral 3 CS (('0.2', '1')) dominates Mistral 3 CS (('0.8', '1')) for this particular prompt.\footnote{When two decoding methods yield identical metric values, they are considered indifferent rather than incomparable. For a detailed distinction between these concepts, see \cite{rodemann2024partial}. For simplicity, we do not differentiate between these cases in the present analysis.} Overall, for each prompt, we derive pairwise comparisons for $6 \text{ models} \times 59 \text{ decoding strategies} = 354$ text continuations, one for each decoding method.

\begin{table*}[!ht]
\centering
\resizebox{1\textwidth}{!}{


\begin{tabular}{p{3.75cm}| p{5.5cm} p{4.5cm} p{4.5cm}}
\hline
\textbf{Characteristic} & \textbf{Extended Bradley-Terry Model} & \textbf{Union-Free Generic Depth} & \textbf{Q*Text}\\ \hline\hline
Considered Information    & Order only  & Order only & Order and metric value  \\ \hline
Methodology & Pairwise comparison & Partial orders & Mean values \\ \hline
Output & Worth Parameter \& Total Order & Partial Order & Mean Values \& Total Order\\
\hline
Results (WikiText-103) & Mistral 3 CS (('0.4', '10')) has the highest worth parameter, while GPT2-XL CS ((’1.0’, ’20’)) has the lowest & The top five models in the Extended Bradley-Terry Model are incomparable, despite the suggested total order & Falcon 2 CS ((’0.8’, ’1’)) has the highest mean and Mistral 3 CS ((’0.2’, ’1’)) the lowest \\
\hline
\end{tabular}

}
\caption{Comparison of the extended Bradley-Terry Model, the ufg-depth and Q*Text (cf. Figure~\ref{fig:introduction}).}
\label{tab:comparison_methods}
\end{table*}

\subsection{Extended Bradley-Terry Model: Theory}\label{sec:bradley_terry}
The \textit{extended Bradley-Terry model} is based on pairwise comparisons \cite{bradley52, davidson70}. It offers a flexible way to rank items while respecting both clear dominance structures and non-dominances (i.e., ties).  Each item $i$, in our situation, decoding method $i$, is assigned a worth parameter $\pi_i$. These worth parameters represent the relative performance/strength of a decoding method in comparison to another decoding method, with all worth parameters summing up to one. The probability that decoding method $i$ is preferred over decoding method $j$ is $ P(i > j) = \pi_i / (\pi_i + \pi_j + \nu \sqrt{\pi_i\pi_j})$.
Here, $\nu$ is a discrimination parameter that reflects the likelihood of a tie, i.e., no preference between the two decoding methods. Based on the estimations, it is possible to conclude that decoding methods with high worth parameters dominate others.

\citet{sinclair82} reformulated the extended Bradley-Terry model as a generalized linear model (GLM) with a Poisson distribution and log link: Let $m_{i >j}$ be the count of times decoding method $i$ outperforms decoding method $j$ and $m_{i \sim j}$ be the number of ties. Then the GLM is given by $\log(m_{i>j}) = \mu_{ij} + \frac{1}{2}\log(\pi_i) - \frac{1}{2}\log(\pi_j)$ and $\log(m_{i\sim j}) = \mu_{ij} + \log(\nu)$ with parameters $\mu_{ij} = \ln{m} - \ln{\left(\sqrt{\pi_i / \pi_j} + \sqrt{\pi_j / \pi_i}\right)}$ and $m$ the total number of pairwise comparisons.

Since it is unlikely that two worth parameters have exactly the same value, the extended Bradley-Terry model yields a total order representing the performance of the decoding methods across all prompts. 

\subsection{Extended Bradley-Terry Model: Experimental Results}
\label{sec:ebt_results}

The extended Bradley-Terry model returns so-called "worth" parameters, which indicate the probability that this decoding method is preferred over the other in a pairwise comparison. When all datasets are considered at once, the method that dominates all other methods according to the extended Bradley-Terry model is Mistral 3 CS (('0.6', '15')). The second-best method is Mistral 3 CS (('0.4', '5')), while the worst method is  GPT2-XL CS (('1.0', '20')). An excerpt of the results, including the case when restricting the analysis to only one dataset, is presented in Table~\ref{tab: bt_wikitext_main}. 

\begin{table}[H]
\centering
\resizebox{0.5\textwidth}{!}{
\begin{tabular}{l S[round-mode=figures,round-precision=2]}
  \hline
Decoding Method & {Estimated worth parameter} \\ 
 \hline\hline
Mistral 3 CS (('0.6', '15')) & 0.04694067374775414552834362780231458600610494613647 \\ 
  Mistral 3 CS (('0.4', '3')) & 0.03744551066886039197845192916247469838708639144897 \\ 
  Mistral 3 CS (('0.8', '3')) & 0.03459659619784605927295118021902453619986772537231 \\ 
  Mistral 3 CS (('0.4', '20')) & 0.02951710689489564845566782480545953148975968360901 \\ 

   \hline
\end{tabular}
}
\caption{Estimated worth parameter of the extended Bradley-Terry model based on WikiText-103 dataset, and the metrics coherence, diversity and perplexity.\label{tab: bt_wikitext_main}}
\end{table}

Note that the total order provided by the extended Bradley-Terry model respects the pairwise dominance structures discussed in Appendix~\ref{a: pairwise_comparison}. As noted above, the extended Bradley-Terry model leads (in almost all cases) to a total order. Hence, it neglects information about incomparabilities. However, the dominance structure provided by the partial orders given by each generation, see Appendix~\ref{a: pairwise_comparison}, already suggests that enforcing a total order (e.g., not allowing incomparability of two decoding methods) may be too strong an assumption. Additionally, the extended Bradley-Terry model relies on further independence assumptions that may not be appropriate for benchmarking purposes \citep{blocher2024comparing}. 

\subsection{Union-Free Generic Depth: Theory}\label{sec:ufg}

\begin{figure}[!ht]
    \centering
    \includegraphics[width=1\linewidth]{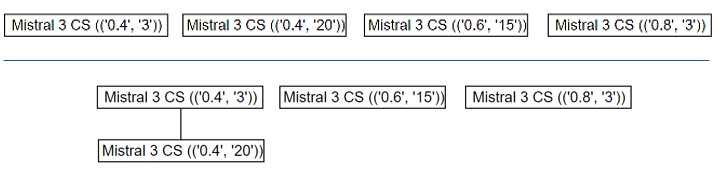}
    \caption{Partial orders with the highest (top) and lowest (bottom) ufg-depths based on Wikitext-103 and the four decoding methods presented in Table \ref{tab: bt_wikitext_main}}
    \label{fig:ufgwikitext-result}
\end{figure}

The \textit{union-free generic (ufg) depth} \citep{rodemann2024partial, blocher2024comparing} directly addresses these concerns by incorporating incomparability information in the estimation itself and avoids any additional independence assumptions. Mathematically, this means that we aim for \textit{partial} rather than \textit{total} orders. Let us look again at a single prompt and the procedure discussed directly before Section~\ref{sec:bradley_terry}. For the extended Bradley-Terry model, we only considered the pairwise comparisons. However, all the pairwise comparisons resulting from one single prompt define a partial order that describes the performance of the decoding methods based on that single prompt. This yields 1,314 partial orders for the WikiText-103 data. For example, in the case where we compare four decoding methods, the two partial orders in Figure~\ref{fig:ufgwikitext-result} correspond to two observations.

The ufg-depth analysis provides a measure for each partial order that indicates how central/typical or outlying/atypical it is. Since each partial order represents the performance of the decoding method, the ufg-depth provides insights into typical and atypical performance structures of the decoding methods. This allows us to identify the \textit{most central} ranking, i.e., the ranking that is most supported by the observed data. To achieve this, the ufg-depth generalizes the well-known simplicial depth from $\mathbb{R}^d$ (which measures centrality by the probability that a point $x$ lies in a randomly drawn $d+1$ simplex \cite{10.1214/aos/1176347507}) to partial orders. This is, \citet{blocher2024comparing} generalize the meaning of "lying in" and "$d+1$ simplex" for $\mathbb{R^d}$, which can be defined by the convex closure operator and the convex sets, to partial orders. Let $\mathcal{P}$ be the set of all partial orders given by the items/decoding methods $m_1, \ldots, m_k$. To transfer the idea of "lying in", \cite{blocher2024comparing} considered the closure operator $\gamma: 2^{\mathcal{P}} \rightarrow 2^{\mathcal{P}}, P \mapsto \{p \in \mathcal{P} \mid \cap_{\tilde{p} \in P} \: \tilde{p} \subseteq p \subseteq \cup_{\tilde{p} \in P }\}$. \citet{blocher2024comparing} showed that $d+1$ simplices in $\mathbb{R}^d$ are those convex sets that are non-trivial, minimal, and not decomposable with respect to the convex closure operator. This is equivalent to consider those sets of partial orders $P = \{p_1, \ldots, p_k\} \in \mathcal{S}\subseteq 2^{\mathcal{P}}$ that satisfy (I) $P \subsetneq \gamma(P)$ and (II) there exists no family $\left(B_i\right)$, with $i \in I$ index, such that $B_i \subseteq P$ and $\gamma(P)=\cup_I \gamma\left(B_i\right)$ (i.e. $P$ cannot be decomposed). The ufg-depth of a partial order $p$ is then the probability that $p$ lies in a randomly drawn $P \in \mathcal{S}$, weighted by the cardinality $P$, see Appendix \ref{a:ufg_technical} for details. For the empirical counterpart, we use the empirical probability measure.

\subsection{Union-Free Generic Depth: Experimental Results}
\label{sec:ufg_results}

Therefore, in the next step, we consider the union-free generic depth approach, which allows for two methods to be incomparable. Furthermore, the ufg-depth considers the entire set of pairwise comparisons for a generation as one observation and does not assume an independence structure between them. Due to the high computational complexity
, we restrict our analysis to the WikiText-103 dataset and compare only the four methods that appear to be the best according to the extended Bradley-Terry model, see Appendix~\ref{a: bradley-terry}: Mistral 3 CS (('0.6', '15')), Mistral 3 CS (('0.4', '3')), Mistral 3 CS (('0.8', '3')) and Mistral 3 CS (('0.4', '20')). 

The highest ufg-depth with a value of 0.977 (thus the one that has the structure most supported by the observation), is the one that shows no dominance structure among the four methods, i.e. the one that concludes that all methods are incomparable to each other, see Figure~\ref{fig:ufgwikitext-result} (top). Roughly speaking, our method reveals that the four decoding methods considered here are incomparable. More formally put, we identify a trivial ranking with no dominance structure as the “central” (in the sense of being the “median”) of the dataset comprising the benchmarking results. This means that such a ranking has most support by the benchmarking results. Our method further finds an “outlier”, i.e., a ranking of methods that has least support by the benchmarking results. In the example at hand, this outlier is a partial ranking that ranks Mistral 3 CS (('0.4', '3')) higher than Mistral 3 CS (('0.4', '20')), see Figure~\ref{fig:ufgwikitext-result} (bottom). This means that, given the benchmarking results, such a ranking of methods is “least central” or “atypical” and therefore based on the benchmarking results with the least supportive structure. 

\section{Scenario 2: Cardinal Assessment} 
\label{sec:collapsing_single_metric}

While multicriteria analysis provides ordinal rankings among decoding methods, many applications require a single unified metric for benchmarking and optimization.

\paragraph{Use Case}

We compute Q*Text scores for over 1.8M text continuations, as described in Table \ref{tab:experimental_setup}, and analyze their performance on a model level, decoding strategy level, and hyperparameter configurations level.

\subsection{Q*Text: Theory}

We propose Q*Text, a text quality metric that integrates coherence, diversity, and perplexity using weighted combinations with Gaussian penalty functions to handle extreme values.

\paragraph{Metric Formulation} Q*Text is defined as:
\begin{equation}
\text{Q*Text} = \frac{\sum_{i=1}^{3} w_i M_i P_i(M_i)}{\sum_{i=1}^{3} w_i}
\end{equation}
where $M_i$ are normalized metrics, $w_i$ are weights, and $P_i(x) = \exp(-\alpha_i (x - \mu_i)^2)$ are Gaussian penalties that discourage extreme values. Parameters $\mu_i$ represent optimal targets while $\alpha_i$ controls penalty strength.

\paragraph{Normalization} We apply inverse normalization to perplexity (lower is better): $M_1 = \frac{p_{\max} - p_i}{p_{\max} - p_{\min}}$, and standard min-max normalization to coherence and diversity (higher is better): $M_j = \frac{m_j - m_{\min}}{m_{\max} - m_{\min}}$ for $j \in \{2,3\}$.

\paragraph{Parameter Optimization} The nine parameters $\theta = \{w_i, \mu_i, \alpha_i\}_{i=1}^{3}$ are optimized via:
\begin{equation}
\theta^* = \text{argmax}_{\theta} \rho_s(\text{Q*Text}(\theta), H)
\end{equation}
where $\rho_s$ is Spearman correlation and $H$ are publicly available human ratings \citep{garces-arias-etal-2025-decoding}. The pseudo-code for the hyperparameter tuning of Q*Text, as well as an interpretation of the resulting values, are presented in Appendix \ref{a:qtext_parameters}, Table \ref{tab:qtext_algorithm}, and Table \ref{tab:qtext_parameters}. Finally, a visualization of the achieved $\rho_s$, highlighting alignments on a decoding strategy level, is illustrated in Appendix \ref{a:qtext_parameters}, Figure \ref{fig:qtext_correlation}.

\subsection{Q*Text: Experimental Results}
\label{sec:qtext_results}

 When analyzing the results we observe the following: For deterministic decoding methods, Q*Text favors balanced hyperparameter choices, particularly CS with moderate penalties ($\alpha$ values of 0.4 or 0.6) and moderate \( k \) values (5, 10, or 15), as shown in Tables \ref{tab: qtext_md_hyper} and \ref{tab: qtext_md_method}. Counterbalancing combinations also perform well, such as low $\alpha$ values (0.2) with high \( k \) values (20 or 50), or high $\alpha$ values (0.8 or 1.0) with moderate \( k \) values (3 or 5). Beam Search (BS) is generally disfavored due to extremely low diversity, indicating Q*Text's capability to penalize \textit{degenerate} text. For stochastic methods, Q*Text prefers diversity-enhancing strategies: temperature sampling with $\tau > 0.7$, top-\( k \) sampling with \( k > 10 \), and nucleus sampling with \( p > 0.8 \).

To illustrate specific results, we sample eight machine-generated continuations of a Wikitext prompt and include the original human text continuation. The text generations are produced by models of different sizes and decoding strategies with varying hyperparameter configurations. The results are presented in Table \ref{tab:case_study_3} and reveal a clear pattern: moderate decoding parameters produce reasonable continuations with scores ranging from 68 to 87, while extreme parameter settings lead to either repetitive or erratic text.

When the degeneration penalty reaches 1.0 with high top-$k$ values, models like GPT2-XL and Qwen 2 generate completely incoherent text with scores near zero. Similarly, Llama 3's beam search produces repetitive, temporally inconsistent content. The human reference completion achieves a score of 87.33, establishing a quality benchmark. Interestingly, GPT2-XL with moderate parameters (degeneration penalty of 0.6, top-$k$ of 5) achieves 86.69, nearly matching human performance despite its relatively small size. These findings align with human evaluator preferences from \cite{garces-arias-etal-2025-decoding}, who favored balanced CS strategies like (('0.4', '10')), (('0.4', '20')), (('0.6', '10')), top-\( k \) sampling with \( k = 10 \) or 50, and nucleus sampling with \( p = 0.9 \). Unbalanced approaches like CS (('1.0', '50')) and BS with moderate beam widths were rated as low quality (Tables \ref{tab:bt_alldatasets2}, \ref{tab: qtext_md_method}). 

\begin{table*}[ht!]
\begin{center}    
\resizebox{1\textwidth}{!}{
    \begin{tabular}{|p{3.5cm}|p{16cm}|p{1.5cm}|}
\hline
\textbf{Decoding Method} & \textbf{Prompt: "Headlam served as Officer Commanding North @-@ Western Area in 1946, and as Director of Training from 1947 to 1950. In 1950 - 51,"} & \textbf{Q*Text}\\ \hline\hline

Mistral 3 CS (0.4, 10) & "he was Chief of Staff of the Malayan Armed Forces. Headlam was promoted to Major General in 1951, and served as GOC Far East Land Forces from 1951 to 1953. He was promoted to Lieutenant General in 1953, and served as Adjutant General from 1953 to 1956.... (text continues)" & 81.62\\ \hline

Mistral 3 CS (0.4, 5) & "he was Commandant of the Royal Military Academy Sandhurst. He was promoted to Major General in 1951, and served as General Officer Commanding 1st Division from 1951 to 1953. Headlam was appointed Chief of the Imperial General Staff in 1953, and held that post until 1956. He was promoted to Field Marshal in 1956, and retired from the Army in 1957 ... (text continues)" & 77.63 \\ \hline

Deepseek CS (0.6, 15) & "he was Chief of Staff (C.O.S.) of the 1st Parachute Brigade, and in 1951 – 52, he was G.O.C. of the 1st Parachute Brigade. He was promoted to Brigadier... (text continues)" & 80.16\\ \hline

Falcon 2 CS (0.8, 1) & "he was the first Commandant of the Staff College, Camberley. He was appointed General Officer Commanding (GOC) 1st Division in 1951, and GOC 3rd Division in 1953. He was appointed Commander-in-Chief (C-in-C) of the British Army of the Rhine (BAOR) in 1955, and retired in 1958... (text continues)" & 68.44\\ \hline

GPT2-XL CS (0.6, 5) & Headlam served as Assistant Chief of Staff, Headquarters U.S. Army (AHCUS) at Fort Leavenworth, Kansas, before being promoted to Brigadier General and assigned to the United States Army Air Forces (USAAF) at Ramstein Air Base, Germany. After the Korean War, Headlam was a consultant to the U.S. Strategic Air Command (SAC) and the National Reconnaissance Office (NRO, which was established in the 1950s to develop intelligence satellites and spy planes). & 86.69\\ \hline

GPT2-XL CS (1.0, 20) & \colorbox{erratic}{"Headless Horsemurdertrud (his nom de veal et re, a translators and transcriph.coastavarism in Latin} \colorbox{erratic}{-Ceux du Ponto-Monahawak at Keremej, New Heythorhampsthembryghtes-sur Meriont Bay of} \colorbox{erratic}{Sulaunkhuyut/Arap
Moheluwempeh is one hundred nautilus middelveth the}... (text continues)" & 0.02\\ \hline

Qwen 2 CS (1.0, 20) & \colorbox{erratic}{" Headelm oversa  Operation Agulhene535af69, to reoccupyo portug 1 st colonial outst a ncesin}
\colorbox{erratic}{AFRicacoun trie sto preva l sepa rabli st mov me s . In th es  on he wro 4he couthoea2 rsh ipp ublicatin}
\colorbox{erratic}{hitherto re t aine o secret}... (text continues)" & 0.02\\ \hline

Llama 3 beam (3) & "he served as Commanding Officer of the 1st Battalion, The Queen's Royal Regiment (West Yorkshire Regiment). In 1951 2013 52, he served as Commanding Officer of the 2nd Battalion, \colorbox{repetitive}{The Queen's Royal Regiment (West Yorkshire Regiment). In 1952 2013 53, he served as Commanding}
 \colorbox{repetitive}{Officer of the 3rd Battalion, The Queen's Royal Regiment (West Yorkshire Regiment). In 1953 2013 54} \colorbox{repetitive}{, he served as Commanding Officer of the 4th Battalion,} ... (text continues)" & 0.02\\ \hline \hline

Human & "he was Director of Operations and Intelligence, and in 1951–54, Commander of the 1st Division, which was the most powerful division in the world. He was appointed Commander-in-Chief of the Army in 1954... (text continues)" & 87.33 \\ \hline
\end{tabular}

}
\captionof{table}{Case Study: Comparison of multiple decoding methods for a prompt from the Wikitext corpus. The first five rows show examples generated by high-ranked methods, while the next three rows display those from low-ranked methods. Human-generated reference text is included for comparison. Degenerate text is highlighted in  \colorbox{repetitive}{purple} while erratic content is highlighted in \colorbox{erratic}{brown}.}
\label{tab:case_study_3}
\end{center}
\end{table*}

\section{Discussion}

First, we examine the extended Bradley-Terry model and the union-free generic depth approach, both of which are based on pairwise comparisons. A first impression can be seen in Appendix~\ref{a: pairwise_comparison}. We observe that out of a total of 124,962 pairwise comparisons among methods, only a very small percentage shows a clear dominance structure. That is, for the majority of the method comparisons, the metrics contradict each other (or imply indifference), e.g., a method is better with respect to coherence but worse with respect to diversity. For a small minority of comparisons, however, methods multilaterally outperform other methods, e.g., method 1 outperforms method 2 for at least 90\% of the generations with respect to all metrics. 

Moving on to Q*Text results, we observe that it shares a preference for larger architectures with the extended Bradley-Terry model, though smaller models like GPT2-XL can outperform modern architectures with balanced decoding strategies (Table \ref{tab: qtext_md_model}).
\\
Agreement analysis between the extended Bradley-Terry model and Q*Text (Appendix~\ref{a: qtext}, Figures \ref{fig: agreement_bt_qtext} and \ref{fig: disagreement_bt_qtext}) highlights discrepancies for less diverse and coherent generations, but good agreement for methods with moderate hyperparameters. The extended Bradley-Terry model does not penalize diversity drops as severely as Q*Text, while both approaches strongly penalize incoherent, low-confidence methods like GPT2-XL with CS ($\alpha = 1.0$, \( k = 20 \)), see Tables \ref{tab: qtext_ld_model}, \ref{tab: qtext_ld_strat} and \ref{tab: qtext_ld_method}.

We now examine the advantages and disadvantages of the three proposed benchmarking methods within our established framework. As highlighted in Section \ref{sec:intro}, benchmarking serves different purposes: Scenario 1 requires only an ordering of decoding methods, while Scenario 2 additionally demands a cardinal assessment of quality. While Scenario 2 naturally encompasses Scenario 1, the ordering focus in Scenario 1 enables the utilization of partial ranking theory, leading to fundamentally different procedures than those based on mean transformations and incorporating concepts such as method incomparability.

Both Scenario 1 methods build upon a data transformation, where metric scores are translated into ordinal values. The \textbf{extended Bradley-Terry Model} offers computational efficiency with $O(n^2m)$ complexity, making it scalable to large numbers of methods and generations. It provides interpretable worth parameters representing estimated preference probabilities and addresses incomparabilities and ties in observed data. However, this approach forces a total order in results, potentially oversimplifying complex dominance structures where methods may genuinely be incomparable. The model assumes independence between pairwise comparisons, which is questionable when comparing methods on fixed datasets, and relies strictly on dominance agreements across all evaluated metrics.

The \textbf{Union-Free Generic Depth} method preserves incomparabilities through partial orderings, providing more realistic representations of method relationships while offering insights into entire performance distribution structures. Unlike the extended Bradley-Terry approach, it does not assume independence between pairwise comparisons, making it more suitable for fixed-dataset evaluations. Nevertheless, this method suffers from computational intensity with worst-case complexity $O(2^m)$, limiting applicability to smaller methods and dataset subsets. The approach is more complex to interpret than traditional rankings and, like the extended Bradley-Terry method, may be overly conservative in establishing dominance relationships.

\textbf{Q*Text} provides cardinal assessment with meaningful score differences, enabling quantification of performance gaps. It incorporates penalization of extreme values to prevent degenerate solutions such as repetitive or erratic text, automatically balances multiple criteria through  mean aggregation, and remains computationally efficient and straightforward to implement. However, the method relies on normalization bounds and penalization parameters that may not generalize across different contexts. By collapsing multiple metrics into a single score, it may obscure important trade-offs between individual metrics and prove less interpretable than separate metric examination, potentially masking insights about specific strengths and weaknesses.

\section{Conclusion}
\label{sec:conclusion}

In this work, we analyze the challenge of evaluating open-ended text generation by introducing a multicriteria benchmarking framework that supports both relative and absolute rankings of decoding methods. We present three complementary approaches—the extended Bradley-Terry model, the union-free generic (ufg) depth, and Q*Text, a unified metric that harmonizes coherence, diversity, and perplexity into a single score. Moreover, we show that our framework captures nuanced trade-offs among metrics and avoids misleading comparisons when methods excel on different criteria.

Extensive experiments involving six large language models, three distinct domains (news, Wikipedia, stories), and over 1.8 million generated continuations demonstrate the practical benefits of our approach. The extended Bradley-Terry model yields interpretable “worth” parameters that reflect overall preference probabilities, while ufg-depth uncovers central and atypical ranking structures, highlighting when decoding methods are genuinely incomparable. Q*Text further enables direct comparison and quantification of performance gaps, revealing that balanced hyperparameter settings outperform extreme configurations and that smaller models can rival larger ones under appropriate decoding choices. Taken together, these contributions provide practitioners and researchers with a more reliable, data-driven basis for selecting and designing decoding methods in open-ended text generation, paving the way for more holistic benchmarking practices.

\section{Key Takeaways and Practical Recommendations}\label{sec:takeaways} Our study revealed that different practical scenarios require different multicriteria benchmark evaluation frameworks. Hence, NLP benchmarking should move beyond a \say{one fits all}-approach. Instead of relying on one single benchmark suite with a pre-specified evaluation method, we recommend that practitioners define the overall aim of benchmarking and evaluation thereof \textit{as precisely as possible}. 

Specifically, we identify two crucial questions to be answered beforehand: 

\begin{enumerate}
    \item Is it sufficient to rank methods, or is metric information about the methods' performances required? (Scenario 1 and 2 in \S\ref{sec:intro})
    \item Does the use case require a total or partial ordering method, i.e., should the evaluation allow for incomparability among some methods, or should it enforce comparability of all methods? (\S\ref{sec:benchmarking})
\end{enumerate}

In case metric information is required and comparability of all methods should be enforced, we recommend our novel aggregation metric Q*Text, see \S\ref{sec:collapsing_single_metric}. If the metric information is not the overall aim, but comparability should still be enforced, we recommend using the Bradley-Terry model, see \S\ref{sec:bradley_terry}. Eventually, if a ranking is required that allows for incomparability, we recommend deploying ufg-depth; see \S\ref{sec:ufg}.

\section*{Limitations}
While our study presents three different benchmarking approaches, this by no means covers the full range of different benchmarking strategies that aim to address the different objectives, i.e., selecting an estimated best method vs. estimating the performance structure of methods. Therefore, this article provides only a glimpse of the complexity and different approaches to multi-metric evaluation. 

Besides this, further limitations merit attention. First, our experiments focused on a limited set of decoding strategies and language models. Alternative methods—such as contrastive decoding \citep{li2023contrastive}, typical sampling \citep{meister2023locally}, and adaptive contrastive search \citep{garces-arias-etal-2024-adaptive}—were not analyzed and may provide insights that refine or challenge our findings.

Secondly, the choice of metrics is a matter of debate. Our reliance on model-dependent metrics, such as coherence, which is measured by an ideally unbiased OPT 2.7B model \cite{zhang2022optopenpretrainedtransformer}, raises questions about their robustness across different models and datasets \citet{he-etal-2023-blind}. Moreover, including further metrics might enhance the robustness and generalizability of our conclusions.

Additionally, while our work focuses on open-ended text generation, the methodologies and insights may also apply to other NLP tasks, such as summarization and machine translation, which present different challenges and evaluation criteria. Applying our framework to these tasks can provide valuable insights into evaluation metrics and benchmarking strategies in broader contexts.

We acknowledge these limitations as avenues for future research. Exploring additional decoding strategies, models, datasets, and metrics will strengthen our approach's validity and adaptability across various language generation tasks, facilitating more nuanced and reliable evaluations.

\section*{Ethics Statement}

We affirm that our research adheres to the \href{https://www.aclweb.org/portal/content/acl-code-ethics}{ACL Ethics Policy}. This work involves the use of publicly available datasets and does not include any personally identifiable information. An ethical concern worth mentioning is the use of language models for text generation, which may produce harmful content, either through intentional misuse by users or unintentionally due to the training data or algorithms. We declare that there are no conflicts of interest that could potentially influence the outcomes, interpretations, or conclusions of this research. All funding sources supporting this study are acknowledged in the acknowledgments section. We have diligently documented our methodology, experiments, and results, and commit to sharing our code, data, and other relevant resources to enhance reproducibility and further advancements in the field.

\section*{Acknowledgments}

Hannah Blocher received financial support via a stipend from Evangelisches Studienwerk Villigst e.V.
Julian Rodemann acknowledges support by the Federal Statistical Office of Germany within the co-operation project "Machine Learning in Official Statistics" as well as by the Bavarian Institute for Digital Transformation (bidt) and the Bavarian Academy of Sciences (BAS) within a graduate scholarship.
Matthias Aßenmacher received funding from the Deutsche Forschungsgemeinschaft (DFG, German Research Foundation) as part of BERD@NFDI, under grant number 460037581. 
\bibliography{custom}

\appendix

\section*{Appendix}
\label{sec:appendix}

\section{Automatic metrics}\label{a:autom_metrics}

\paragraph{Diversity.} This metric aggregates $\mathrm{n}$-gram repetition rates: $$\text{DIV}=\prod_{n=2}^4 \frac{\mid \text { unique } \mathrm{n} \text {-grams }\left(\mathrm{x}_{\text {cont }}\right) \mid}{\mid\text { total } \mathrm{n} \text {-grams }\left(\mathrm{x}_{\text {cont }}\right) \mid}$$ A low diversity score suggests the model suffers from repetition, and a high diversity score means the model-generated text is lexically diverse.

\paragraph{Coherence.} Proposed by \citet{su2022contrastive}, the coherence metric is defined as the averaged log-likelihood of the generated text conditioned on the prompt as
$$
\operatorname{Coherence}(\hat{\boldsymbol{x}}, \boldsymbol{x})=\frac{1}{|\hat{\boldsymbol{x}}|} \sum_{i=1}^{|\hat{\boldsymbol{x}}|} \log p_{\mathcal{M}}\left(\hat{\boldsymbol{x}}_i \mid\left[\boldsymbol{x}: \hat{\boldsymbol{x}}_{<i}\right]\right)
$$

where $\boldsymbol{x}$ and $\hat{\boldsymbol{x}}$ are the prompt and the generated text, respectively; [:] is the concatenation operation and $\mathcal{M}$ is the OPT model (2.7B) \cite{zhang2022optopenpretrainedtransformer}. 

\paragraph{Generation Perplexity.} 
Perplexity \citep{10.1121/1.2016299,   holtzman2019curious} \( P(W) \) of a sequence of words (or tokens) \( W = w_1, w_2, ..., w_N \) is computed as:
\[
P(W) = \exp\left(-\frac{1}{N} \sum_{i=1}^{N} \log p(w_i \mid w_1, ..., w_{i-1}) \right)
\]

Here, \( p(w_i \mid w_1, ..., w_{i-1}) \) is the probability of word \( w_i \) given its preceding context. 

Perplexity measures how well a probabilistic model predicts a sequence of words. Lower perplexity indicates better predictive performance, as the model assigns a higher probability to the actual sequence. It is commonly used to evaluate the quality of language models.

\section{Union-Free Generic Depth}
\label{a:ufg_technical}

\paragraph{General definitions.}
Let $M$ be a set of items/models. $p \subseteq M \times M$ is a partial order (poset) iff $p$ is reflexive (i.e. for all $m \in M,(m, m) \in p$ ), transitive (i.e. $\left.\left(m_1, m_2\right),\left(m_2, m_3\right) \in p \Rightarrow\left(m_1, m_3\right) \in p\right)$ and antisymmetric (i.e. $\left(m_1, m_2\right),\left(m_2, m_1\right) \in p \Rightarrow m_1=m_2$ ).
A closure operator on a set $\Omega$ is a function $\gamma: 2^{\Omega} \rightarrow 2^{\Omega}$ that is extensive (i.e. for all $A \subseteq \Omega$ we have $A \subseteq \gamma(A)$ ), increasing ( $A \subseteq B \subseteq \Omega \Rightarrow \gamma(A) \subseteq \gamma(B)$ ) and idempotent (for all $A \subseteq \Omega, \gamma(A)=\gamma(\gamma(A)))$

\paragraph{Union-free generic depth.}
The definition of the ufg-depth, see \cite{blocher2024comparing}, is analogous to the definition of the simplicial depth on $\mathbb{R}^d$, see \cite{10.1214/aos/1176347507}. Hence, we first have to consider a closure operator $\gamma: 2^{\mathcal{P}} \rightarrow 2^{\mathcal{P}}, P \mapsto\left\{p \in \mathcal{P} \mid \cap_{\tilde{p} \in P} \:\tilde{p} \subseteq p \subseteq \cup_{\tilde{p} \in P}\: \tilde{p}\right\}$. Then a poset $p \in \mathcal{P}$. This is indeed a closure operator and now can be used to generalize the notion of $d+1$ simplices. As described above, we therefore define the set
$$
\mathcal{S}=\{P \subseteq \mathcal{P} \mid \text { Condition }(C 1) \text { and }(C 2) \text { hold }\}
$$

with conditions (C1) $P \subsetneq \gamma(P)$ and (C2) there does not exist a family $(\tilde{P} i) i \in 1, \ldots, \ell$ such that for all $i \in 1, \ldots, \ell \tilde{P} i \subsetneq P$ and $\bigcup_{i \in 1, \ldots, \ell} \gamma(\tilde{P} i)=\gamma(P)$. Note, the (empirical) ufg-depth is given by: Let $p_1, \ldots, p_n \in \mathcal{P}$ be a sample with corresponding empirical probability measure $\nu_n$ (equipped with the power set as $\sigma$-field). Then, the (empirical) union-free generic ( $ ufg$) depth is given by

\begin{tiny}
$$
D_n(p) = \begin{cases}
0, & \text{if } \forall S \in \mathcal{S}: \prod_{\tilde{p} \in S} \nu_n(\tilde{p}) = 0 \\[0.3em]
c_n \displaystyle\sum_{\substack{S \in \mathcal{S} \\ p \in \gamma(S)}} \prod_{\tilde{p} \in S} \nu_n(\tilde{p}), & \text{else}
\end{cases}
$$
\end{tiny}

with $c_n=\left(\sum_{S \in \mathcal{S}} \prod_{\tilde{p} \in S} \nu_n(\tilde{p})\right)^{-1}$. Note that since $\nu_n(p)=0$ if $p \in \mathcal{P}$ is not observed, we can restrict the set $\mathcal{S}$ to $\mathcal{S}_{\text {obs }}=\left\{S \in \mathcal{S} \mid S \subseteq\left\{p_1, \ldots, p_n\right\}\right\}$ consisting only of the observed posets.

Example:
As example consider the four methods Mistral 3 $\mathrm{CS}((0.6, 15))$ (here denoted as $\left.m_1\right)$, Mistral 3 $\mathrm{CS}((0.4, 3))$ (here denoted as $\left.m_2\right)$, Mistral 3 $\mathrm{CS}((0.8, 3))$ (here denoted as $m_3$ ), and Mistral 3 $\mathrm{CS}((0.4, 20))$ (here denoted as $m_4$ ). Assume that the quality metrics provide us with the following four posets:

Let $S = \{(m_i, m_i) \mid i \in \{1,2,3,4\}\}$. Then:
$$
\begin{aligned}
p_1 &= S \cup \{(m_1, m_2)\} \\
p_2 &= S \cup \{(m_1, m_3)\} \\
p_3 &= S \cup \{(m_1, m_2), (m_2, m_3), (m_1, m_3)\} \\
p_4 &= S \cup \{(m_1, m_4)\}
\end{aligned}
$$

Then, with the closure operator above, we get that $p_3 \notin \gamma\left(p_1, p_2\right)$ (note that also incomparabilities are of interest via the union in the definition of the closure operator). The set $\mathcal{S}_{\text{obs}} = \{\{p_1, p_2\}, \{p_1, p_4\}, \{p_2, p_4\}, \{p_3, p_4\},$
$\{p_1, p_2, p_3\}, \{p_1, p_2, p_4\}, \{p_2, p_3, p_4\}\}$. With this, the ufg-depth of $D_n\left(p_1\right)=6 / 7$ and $D_n\left(p_4\right)=5 / 7$. Hence, $p_1$ is more central than $p_4$.

\section{Results of Pairwise Comparisons}\label{a: pairwise_comparison}

The following tables consider the pairwise comparisons of the methods on the generation level, e.g., we count on how many generations one method strictly outperforms another method, compared to \S\ref{sec:bradley_terry}. Since we are comparing $354$ many methods (consisting of model and decoding strategy combination), we have to consider $354 \cdot 353 = 124962$ many pairwise comparisons.

Table~\ref{tab: pairwise_wikitext} collects all pairwise comparisons where Method 1 strictly dominates Method 2 based on all 1314 generations of WikiText-103 and the metrics perplexity, diversity and coherence. Moreover, we can observe that only for 75 of all 124962 pairwise comparisons we have that at least on 90\% of the generations method 1 dominates method 2 strictly. For 30080 pairwise method comparisons, we obtain that method 1 never strictly dominates method 2 (i.e., on every generation, method 2 either dominates method 1 or the three metrics disagree on the dominance structure or are completely equal).

\begin{table}[H]
\centering
\resizebox{0.5\textwidth}{!}{
\begin{tabular}{llr}
  \hline
 Method 1 & Method 2 & count\\ 
  \hline
Mistral 3 CS (('0.2', '1')) & Mistral 3 CS (('0.8', '1')) & 1314 \\ 
Qwen 2 CS (('0.2', '1')) & Qwen 2 CS (('1.0', '1')) & 1314 \\ 
Falcon 2 CS (('0.2', '1')) & Falcon 2 CS (('0.8', '1')) & 1314 \\ 
Falcon 2 CS (('0.2', '1')) & Falcon 2 CS (('1.0', '1')) & 1314 \\ 
Falcon 2 CS (('0.6', '1')) & Falcon 2 CS (('1.0', '1')) & 1314 \\ 
GPT2-XL CS (('0.2', '1')) & GPT2-XL CS (('0.8', '1')) & 1314 \\ 
GPT2-XL CS (('0.4', '1')) & GPT2-XL CS (('0.8', '1')) & 1314 \\ 
GPT2-XL CS (('0.2', '1')) & GPT2-XL CS (('1.0', '1')) & 1314 \\ 
GPT2-XL CS (('0.4', '1')) & GPT2-XL CS (('1.0', '1')) & 1314 \\ 
   \hline
\end{tabular}
}
\caption{All pairwise comparisons of two methods where Method 1 strictly dominates Method 2 based on the three metric perplexity, coherence, and diversity on all 1314 generations of WikiText-103. Count denotes the number of generations where Method 1 strictly dominates Method 2.\label{tab: pairwise_wikitext}}
\end{table}

Table~\ref{tab: pairwise_wikinews} collects all pairwise comparisons where Method 1 strictly dominates Method 2 based on all 2000 generations of Wikinews and the metrics perplexity, diversity, and coherence. Moreover, we can observe that for 878 of all 124,962 pairwise comparisons we have that at least on 90\% of the generations method 1 dominates method 2 strictly. For 25,108 pairwise method comparisons, we obtain that method 1 never strictly dominates method 2 (i.e., on every generation, method 2 either dominates method 1 or the three metrics disagree on the dominance structure or are completely equal). 

\begin{table}[H]
\centering

\resizebox{0.5\textwidth}{!}{
\begin{tabular}{llr}
  \hline
 Method 1 & Method 2 & count\\ 
  \hline
Falcon 2 CS (('0.2', '1')) & Falcon 2 CS (('1.0', '1')) & 2000 \\ 
  Falcon 2 CS (('0.4', '1')) & Falcon 2 CS (('1.0', '1')) & 2000 \\ 
   \hline
\end{tabular}
}
\caption{All pairwise comparisons of two methods where Method 1 strictly dominates Method 2 based on the three metric perplexity, coherence and diversity on all 2000 generations of  Wikinews. Count denotes the number of generations where Method 1 strictly dominates Method 2. \label{tab: pairwise_wikinews}}
\end{table}

Table~\ref{tab: pairwise_books} collects all pairwise comparisons where Method 1 strictly dominates Method 2 based on all 1947 generations of Book and the metrics perplexity, diversity and coherence. Moreover, we can observe that for 546 of all 124962 pairwise comparisons we have that at least on 90\% of the generations method 1 dominates method 2 strictly. For 27947 pairwise method comparisons, we obtain that method 1 never strictly dominates method 2 (i.e. on every generation method 2 either dominates method 1 or the three metrics disagree on the dominance structure or a completely equal).

\begin{table}[H]
\centering
\resizebox{0.5\textwidth}{!}{
\begin{tabular}{llr}
  \hline
 Method 1 & Method 2 & count\\ 
  \hline
Falcon 2 CS (('0.4', '1')) & Falcon 2 CS (('1.0', '1')) & 1947 \\ 
  GPT2-XL CS (('0.4', '15')) & GPT2-XL CS (('1.0', '15')) & 1947 \\ 
   \hline
\end{tabular}
}
\caption{All pairwise comparisons of two methods where Method 1 strictly dominates Method 2 based on the three metric perplexity, coherence and diversity on all 1947 generations of Book. Count denotes the number of generations where Method 1 strictly dominates Method 2. \label{tab: pairwise_books}}
\end{table}

When we merge the three datasets WikiText-103, Wikinews and Book, we consider $1314 + 2000 + 1947 
= 5261$ generations and 124962 pairwise comparisons based on each generation. Comparing the tables~\ref{tab: pairwise_wikitext}, \ref{tab: pairwise_wikinews}, \ref{tab: pairwise_books} we find that there is no pairwise comparison that occurs in each table. Therefore, there is no pair of two methods where method 1 dominates method 2 based on all 5261 generations. With 4601 is the dominance of Mistral 3 CS (('0.8', '10')) over GPT2-XL CS (('1.0', '10'))  the one that occurs most often. For 2990 pairwise comparison at least on 90\% of the generations method 1 dominates method 2 strictly. In 9191 pairwise method comparisons, we obtain that method 1 never strictly dominates method 2 (i.e. on every generation method 2 either dominates method 1 or the three metrics disagree on the dominance structure or a completely equal).

\section{Results of the extended Bradley-Terry model}\label{a: bradley-terry}

In this section, we present the complete result of the extended Bradley-Terry model for all $354$ methods. 

\begin{table}[!ht]
\centering
\resizebox{0.5\textwidth}{!}{
\begin{tabular}{l S[round-mode=figures,round-precision=4]}
  \hline
Method & {Estimated worth parameter} \\ 
 \hline
Mistral3CS0.6\_15 & 0.04694067374775414552834362780231458600610494613647 \\ 
  Mistral3CS0.4\_3 & 0.03744551066886039197845192916247469838708639144897 \\ 
  Mistral3CS0.8\_3 & 0.03459659619784605927295118021902453619986772537231 \\ 
  Mistral3CS0.4\_20 & 0.02951710689489564845566782480545953148975968360901 \\ 
  Mistral3CS0.4\_50 & 0.02673829370438124061748474957767029991373419761658 \\ 
  Mistral3CS0.4\_10 & 0.02198931576528998302344497517424315446987748146057 \\ 
  Mistral3CS0.6\_5 & 0.02143148207667686100386106318183010444045066833496 \\ 
  Qwen2beam50 & 0.01993591751819103960463586133755597984418272972107 \\ 
  Mistral3CS0.6\_20 & 0.01959226140278413560991133124389307340607047080994 \\ 
  Mistral3beam10 & 0.01850553546486544478910651889691507676616311073303 \\ 
  Qwen2beam10 & 0.01808235041824017910738753300847747595980763435364 \\ 
\ldots & \\
  GPT2XLCS0.6\_1 & 0.00005697634689489959001216026757141719372157240286 \\ 
  Falcon2CS1.0\_20 & 0.00005647474586769956825249572318625723710283637047 \\ 
  Mistral3CS1.0\_50 & 0.00005585070054364151096197230184081661263917339966 \\ 
  Falcon2CS1.0\_50 & 0.00005377897546452746387881868606051227743591880426 \\ 
  Mistral3CS1.0\_15 & 0.00005318791435089216459042998907946753206488210708 \\ 
  GPT2XLCS1.0\_1 & 0.00005093752106161674993254490084737540200876537710 \\ 
  GPT2XLCS0.8\_1 & 0.00004712569783000575566935669291623867138696368784 \\ 
  Deepseektemp0.5 & 0.00004617154975830252319323604437428798519249539822 \\ 
  GPT2XLtopk15 & 0.00004077273784737214465129923057773453365371096879 \\ 
  Qwen2CS1.0\_15 & 0.00003622703966772867340256220058947178586095105857 \\ 
  GPT2XLCS1.0\_10 & 0.00003402928365527581100252393908611736605962505564 \\ 
  GPT2XLtopk1 & 0.00003363034106350733980564382541444956586929038167 \\ 
  GPT2XLCS1.0\_20 & 0.00003152902366794530528235768196410049313271883875 \\ 
  GPT2XLtemp0.5 & 0.00002663928436961236458557431006077820256905397400 \\ 
  GPT2XLtopk3 & 0.00002488964926277889070618982991778267432891880162 \\ 
   \hline
\end{tabular}
}
\caption{Estimated worth parameter of the extended Bradley Terry model based on WikiText-103 dataset and the metric coherence, diversity and perplexity. \label{tab: bt_wikitext}}
\end{table}

Note that the higher the estimated worth parameter of the extended Bradley-Terry model, the higher the estimated probability that the method outperforms another method. Hence, the method with the highest worth parameter is, according to the extended Bradley-Terry model, the one that outperforms all others.

\begin{table}[!ht]
\centering
\resizebox{0.5\textwidth}{!}{
\begin{tabular}{l S[round-mode=figures,round-precision=4]}
  \hline
Method & {Estimated worth parameter} \\ 
  \hline
Mistral3CS0.6\_3 & 0.05684612580376772333901058686933538410812616348267 \\ 
  Mistral3CS0.6\_15 & 0.04790914622309408454903234542143763974308967590332 \\ 
  Mistral3CS0.4\_20 & 0.04173233107585151702156878172900178469717502593994 \\ 
  Mistral3CS0.4\_10 & 0.04151904784705818390877496426583093125373125076294 \\ 
  Mistral3CS0.6\_5 & 0.03347410695599519619980455331642588134855031967163 \\ 
  Mistral3CS0.4\_50 & 0.03280097167902953914087049724912503734230995178223 \\ 
  DeepseekCS0.6\_10 & 0.02146468485989244667777597896929364651441574096680 \\ 
  Mistral3CS0.4\_15 & 0.02119850683445620442846291098248912021517753601074 \\ 
  Mistral3CS0.4\_3 & 0.01871866881964552586925520927252364344894886016846 \\ 
  DeepseekCS0.4\_50 & 0.01820406271897827277239656496021780185401439666748 \\ 
  Mistral3CS0.6\_20 & 0.01575772495135861761550444271051674149930477142334 \\ 
  GPT2XLCS0.4\_15 & 0.01552581395605368876078333784107599058188498020172 \\ 
  Mistral3CS0.2\_50 & 0.01508280432494988818059589164022327167913317680359 \\ 
  Mistral3CS0.2\_20 & 0.01386451005337884130608205879298111540265381336212 \\ 
  Mistral3CS0.2\_10 & 0.01266525427607425799414464506753574823960661888123 \\ 
  Mistral3CS0.2\_15 & 0.01231676943599648715865768622279574628919363021851 \\ 
  Mistral3beam5 & 0.01221650293531963485382796363865054445341229438782 \\ 
  Qwen2CS0.6\_5 & 0.01207633440739919704343119377654147683642804622650 \\ 
\ldots & \\ 
  Deepseektemp1 & 0.00007884329880751199490663411184598885483865160495 \\ 
  Deepseektopk3 & 0.00007727586975732274429846946350863845509593375027 \\ 
  Mistral3CS1.0\_5 & 0.00007515570984475984357784522638112889580952469260 \\ 
  GPT2XLtopk20 & 0.00007372338483837541248198249599354880956525448710 \\ 
  Mistral3CS1.0\_10 & 0.00007344322943168122186548274488160359396715648472 \\ 
  Falcon2CS1.0\_50 & 0.00006549324486486910845261383284920952974061947316 \\ 
  Qwen2CS1.0\_15 & 0.00006359548923796265907249641857745814377267379314 \\ 
  GPT2XLtemp1 & 0.00006276976122018022776361673153999731766816694289 \\ 
  Falcon2CS1.0\_15 & 0.00006217193685445184317406308593945141183212399483 \\ 
  Qwen2CS1.0\_10 & 0.00006167572956168872557770888054307079073623754084 \\ 
  Falcon2CS0.8\_5 & 0.00005829604710053048560952984602501203426072606817 \\ 
  GPT2XLtemp0.3 & 0.00005665181267773153025221061218630325129197444767 \\ 
  Falcon2CS1.0\_20 & 0.00005624556365015363811402093752178643626393750310 \\ 
  GPT2XLtopp0.6 & 0.00005571518736143995229655873080432115784788038582 \\ 
  GPT2XLtopk5 & 0.00005211723835054180725429723297636996903747785836 \\ 
  Qwen2CS1.0\_50 & 0.00005211364564949199414817249120801534445490688086 \\ 
  GPT2XLtopp0.7 & 0.00005167290882644772214045414404637313054990954697 \\ 
  GPT2XLtopk3 & 0.00004940536555190984694882483374556159105850383639 \\ 
  GPT2XLCS1.0\_10 & 0.00004934491727118247932602768113241609171382151544 \\ 
  Mistral3CS1.0\_15 & 0.00004752773920582934738055316814531181535130599514 \\ 
  GPT2XLCS1.0\_5 & 0.00004458975768230904080258875099573856459755916148 \\ 
  GPT2XLCS1.0\_20 & 0.00004132810280467744460091567004766943682625424117 \\ 
   \hline
\end{tabular}
}
\caption{Estimated worth parameter of the extended Bradley-Terry model based on Wikinews dataset and the metric coherence, diversity and perplexity. \label{tab: bt_wikinews}}
\end{table}

For reasons of clarity and comprehensibility, we decided to show here only a snippet, but the full result can be easily and fast obtained by the already stored results in GitHub-repository. Table~\ref{tab: bt_wikitext} denotes the worth parameter based on WikiText-103, Table~\ref{tab: bt_wikinews} on Wikinews, Table~\ref{tab: bt_book} on Books and all three datasets combined can be seen in Table~\ref{tab:bt_alldatasets2}. All computations are based on the metrics of perplexity, coherence, and diversity.

\begin{table}[!ht]
\centering
\resizebox{0.5\textwidth}{!}{
\begin{tabular}{l S[round-mode=figures,round-precision=4]}
  \hline
Method & {Estimated worth parameter} \\ 
  \hline
Mistral3CS0.6\_10 & 0.03729170196457783775789707192416244652122259140015 \\ 
  Mistral3CS0.4\_50 & 0.02765856310664721595671267095895018428564071655273 \\ 
  Mistral3CS0.6\_5 & 0.02764552798142631134803437475966347847133874893188 \\ 
  Mistral3CS0.4\_10 & 0.02590282695916623981191584391581272939220070838928 \\ 
  DeepseekCS0.8\_15 & 0.02090843179631032139331736630083469208329916000366 \\ 
  Mistral3CS0.4\_5 & 0.02011703719185768332589070439553324831649661064148 \\ 
  Mistral3CS0.4\_15 & 0.01888515343415263572635076627648231806233525276184 \\ 
  Falcon2CS0.6\_20 & 0.01752855510308615194503545353654772043228149414062 \\ 
  DeepseekCS0.6\_15 & 0.01664459371942486609619038517848821356892585754395 \\ 
  Falcon2CS0.4\_20 & 0.01554863959877443753410108939760903012938797473907 \\ 
  Qwen2CS0.6\_10 & 0.01332183271175595151714077246651868335902690887451 \\ 
  Mistral3beam15 & 0.01237111768271857530077095788101360085420310497284 \\ 
  Qwen2CS0.4\_50 & 0.01218487451617130366832153498535262770019471645355 \\ 
  Qwen2beam5 & 0.01175271030724158068114792996539108571596443653107 \\ 
  Deepseekbeam5 & 0.01174604170705528458085176168879115721210837364197 \\ 
  Mistral3CS0.6\_15 & 0.01095118406217765287535836904453390161506831645966 \\ 
  Mistral3CS0.6\_50 & 0.01087632967263295856452653964652199647389352321625 \\ 
  Falcon2beam15 & 0.01017083098103843154247361013631234527565538883209 \\ 
  Mistral3beam3 & 0.00995040594760476099223112100844446104019880294800 \\ 
  Deepseekbeam15 & 0.00968460062416875641644331551560753723606467247009 \\ 
  Deepseekbeam20 & 0.00952274669027175155178177590187260648235678672791 \\ 
  Mistral3beam20 & 0.00948905214072173763817374947393545880913734436035 \\ 
  Mistral3beam5 & 0.00943927238937268021923276961615556501783430576324 \\ 
\ldots &\\
  DeepseekCS1.0\_50 & 0.00008966777855611055253037128265347632805060129613 \\ 
  Mistral3CS1.0\_15 & 0.00008630467914170437330849405821453501630458049476 \\ 
  GPT2XLCS1.0\_3 & 0.00008525683300025906020871285795337257695791777223 \\ 
  GPT2XLCS0.4\_3 & 0.00008525683300020564969919079079119228481431491673 \\ 
  Qwen2temp0.9 & 0.00008447602119415986666670081994823249260662123561 \\ 
  Mistral3CS1.0\_50 & 0.00008285252330461604264484620774311451896210201085 \\ 
  GPT2XLCS0.4\_5 & 0.00008268205708136934061879691482133125646214466542 \\ 
  GPT2XLtopp0.6 & 0.00007818795559764879650986590942096654544002376497 \\ 
  GPT2XLtopk10 & 0.00007044408307367214629526874780651723995106294751 \\ 
  Falcon2CS1.0\_50 & 0.00006476567332736094288140465957681612962915096432 \\ 
  GPT2XLCS1.0\_5 & 0.00006346211382071562578585655067797688388964161277 \\ 
  GPT2XLCS0.4\_20 & 0.00005905821119543489536914820936530645667517092079 \\ 
  Mistral3CS1.0\_20 & 0.00005601943363400122482620782649043178480496862903 \\ 
  GPT2XLtopk3 & 0.00005049060299638668847547681717280454449792159721 \\ 
  GPT2XLCS1.0\_20 & 0.00004291786826232069323831808116409547437797300518 \\ 
   \hline
   \hline
\end{tabular}
}
\caption{Estimated worth parameter of the extended Bradley-Terry model based on Book dataset and the metric coherence, diversity, and perplexity. \label{tab: bt_book}}
\end{table}

\newpage

\begin{table}[H]
\centering
\resizebox{0.24\textwidth}{!}{
\begin{tabular}{l S[round-mode=figures,round-precision=4]}
  \hline
\multirow{2}{*}{Method} & {Estimated} \\ 
 & {worth parameter} \\ 
  \hline
  \hline
Mistral3CS0.4\_10 & 0.03840793557010361430892331213726720307022333145142 \\ 
Mistral3CS0.4\_5 & 0.03765956236815494462266684649875969626009464263916 \\ 
Mistral3CS0.6\_10 & 0.02173819159864198366505227966172242304310202598572 \\ 
Mistral3CS0.4\_50 & 0.02070821127853808216179309908966388320550322532654 \\ 
Mistral3CS0.6\_15 & 0.01704506832688202494496820804670278448611497879028 \\ 
Mistral3CS0.2\_50 & 0.01649845306850527063002864736063202144578099250793 \\ 
Mistral3CS0.6\_50 & 0.01624336091917789534622151847997884033247828483582 \\ 
Mistral3beam50 & 0.01453010190602210295607310541754486621357500553131 \\ 
Mistral3beam10 & 0.01381660578092141697381656229026702931150794029236 \\ 
Mistral3beam3 & 0.01315480517488667143444214246983392513357102870941 \\ 
Mistral3beam20 & 0.01311640684740963784526979196698448504321277141571 \\ 
Qwen2beam5 & 0.01285665075932509830713978971061806078068912029266 \\ 
Mistral3CS0.4\_1 & 0.01259687265789944006499911921537204761989414691925 \\ 
Mistral3CS0.4\_15 & 0.01162701137550808255793910461761697661131620407104 \\ 
Mistral3beam5 & 0.01154800156717146648877037051761362818069756031036 \\ 
DeepseekCS0.6\_50 & 0.01146320318239432527185339694142385269515216350555 \\ 
Mistral3CS0.6\_20 & 0.01131476677076852579983956559317448409274220466614 \\ 
GPT2XLbeam20 & 0.01087901949185627985261515249248986947350203990936 \\ 
Mistral3CS0.2\_3 & 0.01081200743371361905342808995555969886481761932373 \\ 
Mistral3CS0.2\_15 & 0.01005096549484130186158381548011675477027893066406 \\ 
Qwen2CS0.6\_50 & 0.00999143907722765435563960778608816326595842838287 \\ 
Qwen2beam20 & 0.00996611559455730378065396735109970904886722564697 \\ 
Qwen2CS0.4\_50 & 0.00965872745335187923854114444566221209242939949036 \\ 
Mistral3CS0.2\_10 & 0.00959198264588240721850720404972889809869229793549 \\ 
Qwen2beam3 & 0.00940294902519046502109567597926798043772578239441 \\ 
LLama3beam20 & 0.00899329758576241158896102945163875119760632514954 \\ 
Mistral3CS0.2\_5 & 0.00886838127402570126911474091002673958428204059601 \\ 
Mistral3CS0.6\_5 & 0.00884198718467004395049624321245573810301721096039 \\ 
Mistral3CS0.6\_1 & 0.00850755135707157578162096456253493670374155044556 \\ 
LLama3beam10 & 0.00850479083137893263066864335542049957439303398132 \\ 
LLama3beam3 & 0.00815994538486593951054359763475076761096715927124 \\ 
Qwen2beam50 & 0.00791969910706916829790102951847075019031763076782 \\ 
LLama3beam5 & 0.00763580408619234964062716386479223729111254215240 \\ 
Qwen2CS0.4\_20 & 0.00761302607576399420485913083211926277726888656616 \\ 
Qwen2beam15 & 0.00744512933797864158436130566087740589864552021027 \\ 
Falcon2CS0.6\_50 & 0.00736408333218464394159807895334779459517449140549 \\ 
Qwen2beam10 & 0.00730726198733271364166297345832390419673174619675 \\ 
Mistral3CS0.4\_3 & 0.00724176029325422718901927865431389363948255777359 \\ 
Qwen2CS0.4\_15 & 0.00723628711890984994969944210652101901359856128693 \\ 
GPT2XLCS0.6\_10 & 0.00711343685346400769059505009295207855757325887680 \\ 
Mistral3CS0.8\_5 & 0.00678110758849108027912500062939216149970889091492 \\ 
Falcon2beam15 & 0.00652577783768665004721087896655262738931924104691 \\ 
LLama3beam50 & 0.00624626566939979512016467566581923165358603000641 \\ 
LLama3beam15 & 0.00617542600809433545050453062685846816748380661011 \\ 
Mistral3beam15 & 0.00609687813718103738258591306475864257663488388062 \\ 
Deepseekbeam10 & 0.00607251094733232948258416783460234000813215970993 \\ 
Mistral3CS0.2\_1 & 0.00601451519858165264964533491820475319400429725647 \\ 
Falcon2beam5 & 0.00589795601762234996540135156806172744836658239365 \\ 
DeepseekCS0.8\_15 & 0.00578940322321826010554257635476460563950240612030 \\ 
Qwen2CS0.4\_5 & 0.00571718663085621272429426298344878887291997671127 \\ 
Falcon2CS0.4\_50 & 0.00540997138991246556000147549525536305736750364304 \\ 
Qwen2CS0.2\_1 & 0.00538217521480945527284633911335731681901961565018 \\ 
Deepseekbeam3 & 0.00532804987981470683233720109228670480661094188690 \\ 
Qwen2CS0.2\_50 & 0.00518913827679240782786429520001547643914818763733 \\ 
Mistral3topp0.7 & 0.00494273700694439904745891567472426686435937881470 \\ 
Falcon2CS0.4\_20 & 0.00492427885778739214683863423260845593176782131195 \\ 
Qwen2CS0.2\_15 & 0.00479058417486692787040647445451213570777326822281 \\ 
Qwen2CS0.6\_20 & 0.00477920639581254184935499296216221409849822521210 \\ 
DeepseekCS0.4\_20 & 0.00473047566013182121252844680725502257701009511948 \\ 
GPT2XLbeam5 & 0.00472354697346990051731729920447833137586712837219 \\ 
Mistral3CS0.2\_20 & 0.00470941382939106967936426784149261948186904191971 \\ 
Falcon2CS0.2\_20 & 0.00465810258971982235620679091425699880346655845642 \\ 
DeepseekCS0.8\_10 & 0.00463670848877492745160333598164470458868891000748 \\ 
Falcon2beam50 & 0.00458850962459668269832757658832633751444518566132 \\ 
Deepseekbeam50 & 0.00451256198267279036440147166331371408887207508087 \\ 
Falcon2beam3 & 0.00443540704962581879694516473477960971649736166000 \\ 
Falcon2beam10 & 0.00434473847973430166158337328852212522178888320923 \\ 
Falcon2CS0.4\_3 & 0.00432099306570987284209506285037605266552418470383 \\ 
Deepseekbeam15 & 0.00429819034292664835933406308754456404130905866623 \\ 
Falcon2CS0.4\_15 & 0.00428015253692712923511143330301820242311805486679 \\ 
Falcon2CS0.4\_10 & 0.00421163607677808599777380393902603827882558107376 \\ 
Deepseekbeam5 & 0.00412479234214195404628622299014750751666724681854 \\ 
DeepseekCS0.6\_15 & 0.00407906762456166759739373262050321500282734632492 \\ 
Falcon2CS0.6\_20 & 0.00394866044739210046710109480727624031715095043182 \\ 
Falcon2CS0.4\_1 & 0.00389328824811408695863446105533967056544497609138 \\ 
Qwen2CS0.2\_5 & 0.00389046217006713606831325158452727919211611151695 \\ 
Mistral3CS0.4\_20 & 0.00388030240212250843861996330019792367238551378250 \\ 
Qwen2CS0.2\_20 & 0.00374618937464818533428223545911350811365991830826 \\ 
Falcon2CS0.6\_3 & 0.00374385056147143744156857714244779344880953431129 \\ 
Falcon2CS0.6\_10 & 0.00369049236685878301322816241736290976405143737793 \\ 
Falcon2CS0.2\_50 & 0.00365139250664532925491623238656302419258281588554 \\ 
Falcon2CS0.6\_15 & 0.00364263217386410478523806055761724564945325255394 \\ 
Falcon2CS0.2\_15 & 0.00356243329650930348750925702461245236918330192566 \\ 
DeepseekCS0.2\_10 & 0.00352693852401568877508175603452400537207722663879 \\ 
Falcon2CS0.2\_10 & 0.00351375822380327025931667783709144714521244168282 \\ 
DeepseekCS0.2\_20 & 0.00350692469283286119979825556924879492726176977158 \\ 

  \hline
\end{tabular}
}
\end{table}

\begin{table}[H]
\centering
\resizebox{0.24\textwidth}{!}{
\begin{tabular}{l S[round-mode=figures,round-precision=4]}
  \hline
\multirow{2}{*}{Method} & {Estimated} \\ 
 & {worth parameter} \\ 
  \hline
  \hline

Qwen2CS0.2\_10 & 0.00350410031133972744640803220761426928220316767693 \\ 
Falcon2CS0.2\_3 & 0.00345121950463527773708682921949275623774155974388 \\ 
Falcon2beam20 & 0.00339429572697707085393581394328066380694508552551 \\ 
GPT2XLCS0.6\_5 & 0.00331857643778114917015864548943682166282087564468 \\ 
DeepseekCS0.2\_15 & 0.00325706076977763355276285217598797316895797848701 \\ 
GPT2XLCS0.4\_50 & 0.00322474066426160386675259239552815415663644671440 \\ 
Falcon2CS0.2\_1 & 0.00317387709519370607783028681581072305561974644661 \\ 
DeepseekCS0.2\_3 & 0.00315334692835002837474345227519734180532395839691 \\ 
Deepseekbeam20 & 0.00312251001857651163881590683502054162090644240379 \\ 
Falcon2CS0.4\_5 & 0.00309980742214592064689981931735474063316360116005 \\ 
DeepseekCS0.4\_10 & 0.00293371805639143978447491711847305850824341177940 \\ 
Falcon2CS0.2\_5 & 0.00291880125463333346333416606910304835764691233635 \\ 
Qwen2CS0.4\_1 & 0.00278190629569124948591851698154187033651396632195 \\ 
DeepseekCS0.4\_50 & 0.00263616075638965861177576144314116390887647867203 \\ 
Qwen2CS0.6\_10 & 0.00262668673257979374061066302203926170477643609047 \\ 
Qwen2CS0.8\_1 & 0.00260517128741147860684912451745276484871283173561 \\ 
GPT2XLCS0.6\_50 & 0.00254912072671362134254624542961664701579138636589 \\ 
GPT2XLbeam10 & 0.00252215505036754031667478592737552389735355973244 \\ 
GPT2XLbeam3 & 0.00248105231046832691230163980833367531886324286461 \\ 
Qwen2CS0.4\_10 & 0.00248000722748790503394133999393034173408523201942 \\ 
DeepseekCS0.2\_1 & 0.00247753543764274355293286689061460492666810750961 \\ 
Mistral3CS0.6\_3 & 0.00246815585076946624951133024694627238204702734947 \\ 
GPT2XLbeam15 & 0.00245656254528779649515257688108249567449092864990 \\ 
GPT2XLCS0.6\_50 & 0.00254912072671362134254624542961664701579138636589 \\ 
GPT2XLbeam10 & 0.00252215505036754031667478592737552389735355973244 \\ 
GPT2XLbeam3 & 0.00248105231046832691230163980833367531886324286461 \\ 
Qwen2CS0.4\_10 & 0.00248000722748790503394133999393034173408523201942 \\ 
DeepseekCS0.2\_1 & 0.00247753543764274355293286689061460492666810750961 \\ 
Mistral3CS0.6\_3 & 0.00246815585076946624951133024694627238204702734947 \\ 
GPT2XLbeam15 & 0.00245656254528779649515257688108249567449092864990 \\ 
GPT2XLbeam50 & 0.00245324931627677149392763311652743141166865825653 \\ 
Mistral3topp0.8 & 0.00238701614210672444227334665356465848162770271301 \\ 
Qwen2CS0.6\_5 & 0.00232887733228399809032316269963303057011216878891 \\ 
Falcon2CS0.6\_5 & 0.00231712301287552392345392426875605451641604304314 \\ 
Qwen2CS0.4\_3 & 0.00230715138935352935892209025325882976176217198372 \\ 
DeepseekCS0.2\_50 & 0.00224683101365222639977936580635287100449204444885 \\ 
Mistral3topp0.6 & 0.00219330713908374953444169186411727423546835780144 \\ 
Qwen2CS0.6\_1 & 0.00217511902580839111207011704607339197536930441856 \\ 
Qwen2CS0.2\_3 & 0.00216118472952793721358899503570683009456843137741 \\ 
Falcon2CS0.8\_10 & 0.00213216539276257820503812645540619996609166264534 \\ 
Falcon2CS0.8\_20 & 0.00211836405864725347072963224093200551578775048256 \\ 
DeepseekCS0.6\_1 & 0.00209438018675693442272045530216928455047309398651 \\ 
Mistral3CS0.8\_10 & 0.00204637553966052982215573230462268838891759514809 \\ 
DeepseekCS0.4\_1 & 0.00203358322625844713943399355571273190435022115707 \\ 
DeepseekCS0.8\_20 & 0.00201947924960769282717865102938503696350380778313 \\ 
DeepseekCS0.8\_3 & 0.00195560564210130616996652364036890503484755754471 \\ 
GPT2XLCS0.6\_20 & 0.00195012219888833428400876091046711735543794929981 \\ 
LLama3temp0.9 & 0.00192151481710344006550994500059914571465924382210 \\ 
GPT2XLCS0.2\_50 & 0.00192068833020464643443481200790756702190265059471 \\ 
DeepseekCS0.4\_15 & 0.00188584312642831150759470393296624024515040218830 \\ 
GPT2XLCS0.8\_1 & 0.00187439693765616035951715456775446000392548739910 \\ 
Falcon2CS0.6\_1 & 0.00185207367863864102996984684068593196570873260498 \\ 
DeepseekCS1.0\_20 & 0.00184503335226524306338091729884354208479635417461 \\ 
GPT2XLCS0.6\_1 & 0.00183910205171632936370362809697098782635293900967 \\ 
GPT2XLCS0.8\_15 & 0.00181615789757561571331201388801446228171698749065 \\ 
GPT2XLCS0.4\_10 & 0.00180002838449196188135015272990813173237256705761 \\ 
Mistral3CS0.8\_1 & 0.00178472882134247967135520784864866072894074022770 \\ 
GPT2XLCS0.6\_3 & 0.00176540340778911521678329332729617817676626145840 \\ 
Falcon2temp0.1 & 0.00176265898369564236701501069859432391240261495113 \\ 
Mistral3temp0.5 & 0.00176138979675304302267846434659759324858896434307 \\ 
DeepseekCS0.6\_5 & 0.00173834237103979213218130084328549855854362249374 \\ 
LLama3CS1.0\_15 & 0.00170266532514298861260282347274142011883668601513 \\ 
LLama3CS0.2\_15 & 0.00168013719978836353723117280623000624473206698895 \\ 
GPT2XLCS0.2\_5 & 0.00165952510660342207257644808748864306835457682610 \\ 
Deepseektopp0.6 & 0.00165557551446148453963558289103730203351005911827 \\ 
Qwen2topp0.6 & 0.00165371159568247468388269005856727744685485959053 \\ 
LLama3topk15 & 0.00161921846400833301798938279603135015349835157394 \\ 
GPT2XLCS0.8\_5 & 0.00160282512150825072298754303545820221188478171825 \\ 
GPT2XLtemp1 & 0.00158107316571760316524508649393965242779813706875 \\ 
Mistral3temp0.3 & 0.00155699367265110127402094608584093293757177889347 \\ 
GPT2XLCS0.2\_10 & 0.00153551522211304919060537166330959735205397009850 \\ 
GPT2XLCS0.2\_15 & 0.00151412964903740077278471165556084088166244328022 \\ 
LLama3temp0.3 & 0.00149825541958130653390213193887348097632639110088 \\ 
Falcon2topp0.9 & 0.00147733703570968109668182233207289755227975547314 \\ 
DeepseekCS0.6\_10 & 0.00146925050398758394107345814916243398329243063927 \\ 
LLama3temp0.7 & 0.00146368589045604586412752112778434820938855409622 \\ 
GPT2XLCS0.2\_3 & 0.00145632689528450189053732088950710021890699863434 \\ 
Falcon2topk20 & 0.00145253487804555413316631717890459185582585632801 \\ 
LLama3CS0.2\_5 & 0.00145181970507448992145638211326286182156763970852 \\ 
Mistral3topk15 & 0.00144533599160124801019655649270134745165705680847 \\ 
Mistral3temp0.9 & 0.00142903447768662271270823094226898319902829825878 \\ 
Qwen2topp0.95 & 0.00141923428409681823963994773407648608554154634476 \\ 
LLama3CS0.6\_5 & 0.00140768448210307748659497750765012824558652937412 \\ 
LLama3CS0.8\_5 & 0.00140310296583640611682575194407718299771659076214 \\ 
Mistral3topk5 & 0.00139691614632485847959098013149059624993242323399 \\ 
GPT2XLCS0.4\_15 & 0.00138848126823775991529885676101230274070985615253 \\ 
  \hline
\end{tabular}
}
\end{table}

\begin{table}[H]
\centering
\resizebox{0.24\textwidth}{!}{
\begin{tabular}{l S[round-mode=figures,round-precision=4]}
  \hline
\multirow{2}{*}{Method} & {Estimated} \\ 
 & {worth parameter} \\ 
  \hline
  \hline
Qwen2topk1 & 0.00135191276435400894138927352372547829872928559780 \\ 
Deepseektemp0.7 & 0.00134145389263769340866538914980310437385924160480 \\ 
LLama3CS0.4\_5 & 0.00130016100135360666151251862743265519384294748306 \\ 
Qwen2CS0.6\_3 & 0.00129598634970833135986179041054811023059301078320 \\ 
Falcon2topp0.7 & 0.00129141359527988673454312884558703444781713187695 \\ 
Mistral3topk50 & 0.00129029164484332375430197803467535777599550783634 \\ 
Qwen2CS0.6\_15 & 0.00127905205249944134947293061799200586392544209957 \\ 
GPT2XLCS0.2\_1 & 0.00126844411938711832885107178015005047200247645378 \\ 
GPT2XLCS0.2\_20 & 0.00125299843347637974988462161718416609801352024078 \\ 
LLama3CS0.8\_50 & 0.00124498671701313897569596278458448068704456090927 \\ 
Falcon2temp0.3 & 0.00122235919337801814468780214895105018513277173042 \\ 
DeepseekCS0.8\_50 & 0.00120474402121656669513438675522820631158538162708 \\ 
LLama3CS1.0\_5 & 0.00120449998938870436217263293343648911104537546635 \\ 
Mistral3topp0.9 & 0.00119244887683182506157852920125606033252552151680 \\ 
Qwen2topk15 & 0.00118596349860753628308041029981723113451153039932 \\ 
Falcon2temp1 & 0.00117653281860917663728671733736064197728410363197 \\ 
LLama3CS0.8\_15 & 0.00117272785271453237540550951223394804401323199272 \\ 
LLama3CS0.4\_50 & 0.00116652111342125676916869903010365305817686021328 \\ 
Qwen2temp0.1 & 0.00116180843543374663628942045079384115524590015411 \\ 
GPT2XLCS0.6\_15 & 0.00116172445860343717324858481276805832749232649803 \\ 
DeepseekCS0.4\_3 & 0.00115743317787229526433234738647115591447800397873 \\ 
Falcon2topk3 & 0.00114856746649245476839251711709266601246781647205 \\ 
Falcon2CS0.8\_3 & 0.00114118886422363862744033813356736573041416704655 \\ 
DeepseekCS1.0\_10 & 0.00111280294768279506070762785441274900222197175026 \\ 
LLama3temp0.5 & 0.00111187156925323586513509876994021396967582404613 \\ 
Falcon2topk1 & 0.00110698708013670914923620713921081915032118558884 \\ 
LLama3CS1.0\_50 & 0.00110470464096503622787592835408077007741667330265 \\ 
DeepseekCS0.2\_5 & 0.00108936646016443913764781825648242374882102012634 \\ 
GPT2XLCS0.4\_1 & 0.00108586804732668427107833419853477607830427587032 \\ 
LLama3CS0.6\_50 & 0.00107033873366607157687213636165779462317004799843 \\ 
Falcon2topp0.8 & 0.00106550559948202653096083203365651570493355393410 \\ 
LLama3topp0.9 & 0.00106289844177680226078308578507858328521251678467 \\ 
LLama3CS0.6\_10 & 0.00098200196246093964663748110410779190715402364731 \\ 
Qwen2topp0.7 & 0.00096969564570667141859688920035864612145815044641 \\ 
LLama3CS0.4\_15 & 0.00096592073008356683916364460174008854664862155914 \\ 
LLama3CS0.2\_20 & 0.00096410822683672917510905309157465126190800219774 \\ 
LLama3CS0.8\_10 & 0.00095957210790712464936608183663224735937546938658 \\ 
LLama3CS0.4\_1 & 0.00095921459755446815496210399842880178766790777445 \\ 
GPT2XLCS0.4\_5 & 0.00095837637827738070619987276188567193457856774330 \\ 
LLama3CS0.8\_20 & 0.00095801934092746145014973802034319305676035583019 \\ 
Deepseektopk20 & 0.00094630249391429693394534039541099446068983525038 \\ 
Mistral3topk20 & 0.00092710767387789525667551870924398826900869607925 \\ 
LLama3CS0.6\_20 & 0.00091536144838414985717978122892191095161251723766 \\ 
Mistral3topk1 & 0.00090327125769320800504486035364948293135967105627 \\ 
LLama3CS0.6\_3 & 0.00090292319167467615954186799243075256526935845613 \\ 
LLama3CS0.2\_1 & 0.00089982274286870100703239483408424348453991115093 \\ 
Mistral3topk10 & 0.00089338841477156264152953424328984510793816298246 \\ 
LLama3CS1.0\_1 & 0.00088981540963261305816206503394028004549909383059 \\ 
Falcon2CS0.8\_50 & 0.00088723110410260418515204783318495174171403050423 \\ 
LLama3CS0.8\_3 & 0.00087976460222938918879359571079135093896184116602 \\ 
LLama3CS0.8\_1 & 0.00087543935317600123958980962157738758833147585392 \\ 
Falcon2topk50 & 0.00087272596138171390587473519317995851451996713877 \\ 
Qwen2CS1.0\_1 & 0.00087102539481277317569979379285882714611943811178 \\ 
LLama3CS0.2\_3 & 0.00087012071257540986226469703623820350912865251303 \\ 
LLama3CS1.0\_10 & 0.00086826495470703119773647760126777939149178564548 \\ 
LLama3CS1.0\_3 & 0.00086759792717829172480070720752109991735778748989 \\ 
LLama3CS1.0\_20 & 0.00085546660803597479854482044103747284680139273405 \\ 
Qwen2CS0.8\_15 & 0.00085508970953083606077183009830378068727441132069 \\ 
Qwen2CS1.0\_15 & 0.00085352325228810394831813201932391166337765753269 \\ 
LLama3CS0.2\_10 & 0.00085095565633855211121194805912182346219196915627 \\ 
Qwen2topp0.8 & 0.00084898719668699551554880144976777955889701843262 \\ 
Qwen2temp0.3 & 0.00084889071211262693509441712436114357842598110437 \\ 
LLama3topk5 & 0.00084849283868029313528114077058717157342471182346 \\ 
Qwen2topk50 & 0.00082433190542085155345930491677108875592239201069 \\ 
GPT2XLCS0.4\_3 & 0.00082368967661281488300273334601797614595852792263 \\ 
LLama3temp0.1 & 0.00080168728127494423736643103239885022048838436604 \\ 
Mistral3CS1.0\_20 & 0.00078382137457053808159146424117125206976197659969 \\ 
LLama3CS0.6\_1 & 0.00077874231727603196364512205818186885153409093618 \\ 
Qwen2temp0.7 & 0.00077592649744568839825092920747806601866614073515 \\ 
Deepseektemp1 & 0.00076951517009884345173892361557932417781557887793 \\ 
Falcon2topk10 & 0.00074189141111371756116127373914537201926577836275 \\ 
Deepseektopk3 & 0.00073959517129882810463487041729990778549108654261 \\ 
Deepseektopk10 & 0.00072969087982660920700672679828358013764955103397 \\ 
Mistral3CS1.0\_5 & 0.00072887386951013240568575390909700217889621853828 \\ 
DeepseekCS1.0\_3 & 0.00070898237125087133596312849448395354556851089001 \\ 
Qwen2CS0.8\_50 & 0.00070870529278632828865808201967979584878776222467 \\ 
Mistral3CS0.8\_20 & 0.00070056586713007607501724205434356917976401746273 \\ 
Falcon2CS0.8\_15 & 0.00069788561069875792476635556482733591110445559025 \\ 
LLama3CS0.2\_50 & 0.00069134547974144730780182710461190254136454313993 \\ 
GPT2XLCS0.4\_20 & 0.00069035499244995014028747348788783710915595293045 \\ 
LLama3topk50 & 0.00067703858813675016261462991096209407260175794363 \\ 
Qwen2temp1 & 0.00066889540895948939973336688780136682908050715923 \\ 
Falcon2topp0.95 & 0.00064699079923311432569615897492099065857473760843 \\ 
LLama3CS0.4\_20 & 0.00064547385095997569359055345827869132335763424635 \\ 
LLama3topk20 & 0.00064192918811095190430998647102001086750533431768 \\ 
LLama3topk3 & 0.00064142459981405056373277018266776394739281386137 \\ 
Falcon2topp0.6 & 0.00063952534636457433747658285483339568600058555603 \\ 
LLama3topp0.8 & 0.00063891716260335202921122332853087755211163312197 \\ 
Qwen2CS0.8\_20 & 0.00063087875294048922653750155475904648483265191317 \\ 
Mistral3temp0.1 & 0.00062701882421098192790104297600350946595426648855 \\ 
LLama3topk1 & 0.00062534817535268545987681143571990105556324124336 \\ 
LLama3CS0.4\_3 & 0.00062395695233132258007008319466990542423445731401 \\ 
Falcon2CS1.0\_3 & 0.00062137277233088926507026394929766865971032530069 \\ 
LLama3CS0.6\_15 & 0.00061628295093444545107996823674056940944865345955 \\ 
Qwen2topk20 & 0.00061577572550336175192381427123677894996944814920 \\ 
  \hline
\end{tabular}
}
\end{table}

\begin{table}[H]
\centering
\resizebox{0.235\textwidth}{!}{
\begin{tabular}{l S[round-mode=figures,round-precision=4]}
  \hline
\multirow{2}{*}{Method} & {Estimated} \\ 
 & {worth parameter} \\ 
  \hline
  \hline

GPT2XLCS0.8\_3 & 0.00061269187358480893534906641662018955685198307037 \\ 
Mistral3CS0.8\_50 & 0.00060894925041347230682681201940908977121580392122 \\ 
Deepseektopk15 & 0.00060629419618812776127808605863833690818864852190 \\ 
Falcon2CS1.0\_5 & 0.00060549775659098732313867508025850838748738169670 \\ 
DeepseekCS1.0\_15 & 0.00060530711933960094223161618742778955493122339249 \\ 
DeepseekCS0.8\_5 & 0.00060000505579593864206866538069107264163903892040 \\ 
DeepseekCS0.6\_20 & 0.00059489382536201504576844900640253399615176022053 \\ 
GPT2XLtopp0.95 & 0.00058769610189011848612461452745492351823486387730 \\ 
Qwen2topp0.9 & 0.00058664741385072390923716145749722272739745676517 \\ 
LLama3CS0.4\_10 & 0.00057674747371252641208599909461440802260767668486 \\ 
Deepseektemp0.3 & 0.00057333934830933804428082511961406453337986022234 \\ 
LLama3topk10 & 0.00057170603738173513593684438305331241281237453222 \\ 
DeepseekCS0.6\_3 & 0.00055857996118351040809590912772364390548318624496 \\ 
GPT2XLCS0.8\_10 & 0.00055410926482804632314804749171344155911356210709 \\ 
Mistral3CS1.0\_1 & 0.00054582041786915575648508891504206985700875520706 \\ 
Deepseektopp0.7 & 0.00054484212687568533105275658812161054811440408230 \\ 
LLama3topp0.95 & 0.00053902909102668088159315828988837893120944499969 \\ 
Mistral3CS0.8\_15 & 0.00053060687162623059159177740795598765544127672911 \\ 
GPT2XLtopk1 & 0.00052973105548507148053472004534114603302441537380 \\ 
Mistral3topk3 & 0.00052065947343785210377903327483295470301527529955 \\ 
Falcon2CS0.8\_5 & 0.00052044335863967567355758969682710812776349484921 \\ 
Falcon2CS1.0\_10 & 0.00051380186708916721332468524607861581898760050535 \\ 
Qwen2temp0.5 & 0.00050544216155160612106989059810757680679671466351 \\ 
GPT2XLtopp0.7 & 0.00049987241549829261946791270077028457308188080788 \\ 
Qwen2CS0.8\_10 & 0.00048752824614378518922400140844786164961988106370 \\ 
Qwen2topk5 & 0.00048567812641822887941706055059398750017862766981 \\ 
GPT2XLCS0.8\_20 & 0.00048041160549572108594273966986065715900622308254 \\ 
Mistral3topp0.95 & 0.00046714088978155762733532330166497104073641821742 \\ 
DeepseekCS0.4\_5 & 0.00045224454002698030253074024109594120091060176492 \\ 
DeepseekCS1.0\_5 & 0.00044043542235632654208848935084574804932344704866 \\ 
Falcon2CS1.0\_20 & 0.00043747038981853804233412841284689420717768371105 \\ 
Qwen2topk10 & 0.00043645890307528596675432086549051291513023898005 \\ 
Mistral3temp1 & 0.00043503727499504126711299401009114262706134468317 \\ 
GPT2XLtopk5 & 0.00042603075965480372748819815065246530139120295644 \\ 
Qwen2topk3 & 0.00042132976589705033437088799175285203091334551573 \\ 
Qwen2CS0.8\_5 & 0.00041905096527107652416149763396902017120737582445 \\ 
GPT2XLtemp0.3 & 0.00041404611827043408169707339183673866500612348318 \\ 
LLama3temp1 & 0.00040988555553102306227772344549009631009539589286 \\ 
Falcon2temp0.7 & 0.00039160443941076280779944873700060270493850111961 \\ 
Falcon2topk15 & 0.00038809524071708193177290824316116868430981412530 \\ 
Falcon2temp0.5 & 0.00038559692125231573821275699032185002579353749752 \\ 
LLama3topp0.6 & 0.00038027674789149138855764986644203418109100311995 \\ 
LLama3topp0.7 & 0.00037835520276987178074634687874322480638511478901 \\ 
Falcon2topk5 & 0.00037599809052900746812742038827082069474272429943 \\ 
Deepseektemp0.5 & 0.00035448731737414742584343918530009887035703286529 \\ 
GPT2XLtemp0.7 & 0.00035207435991842586393146818934951625124085694551 \\ 
Mistral3CS0.8\_3 & 0.00034803875490391640015394592033715071011101827025 \\ 
Deepseektopp0.95 & 0.00034294139508813991984101376964133578439941629767 \\ 
Qwen2CS0.8\_3 & 0.00033910282818424035064519550530803826404735445976 \\ 
Deepseektopk50 & 0.00033854074311681951140601265493046412302646785975 \\ 
Deepseektopp0.9 & 0.00033475734721354115592042988147625237616011872888 \\ 
Falcon2CS0.8\_1 & 0.00033021859744564483829867840292138225777307525277 \\ 
Deepseektopp0.8 & 0.00032953225380896831054472473354621797625441104174 \\ 
GPT2XLtopk50 & 0.00032906620179489582626169985601904954819474369287 \\ 
GPT2XLtopp0.9 & 0.00032873437342544357606111127267922711325809359550 \\ 
GPT2XLtemp0.9 & 0.00031492982663750836939270394410073095059487968683 \\ 
Qwen2CS1.0\_3 & 0.00031088811649074884601445800669239361013751477003 \\ 
DeepseekCS0.8\_1 & 0.00030564735523578461830196406623372240574099123478 \\ 
Mistral3temp0.7 & 0.00029779789423327389071352588878482947620796039701 \\ 
GPT2XLCS1.0\_3 & 0.00029748962957066102255967998679864194855326786637 \\ 
GPT2XLtopk3 & 0.00029225560683055206183042495915458403032971546054 \\ 
GPT2XLCS1.0\_1 & 0.00028732232131022865211089634129848491284064948559 \\ 
Qwen2temp0.9 & 0.00028530195340966671077587157867583300685510039330 \\ 
Deepseektopk5 & 0.00028197649037294832781094799933896410948364064097 \\ 
Mistral3CS1.0\_15 & 0.00027448090875735112323038911164019282296067103744 \\ 
Mistral3CS1.0\_10 & 0.00026839361244045101782818019131582332192920148373 \\ 
Falcon2CS1.0\_15 & 0.00026508682696197035343066428225711206323467195034 \\ 
Mistral3CS1.0\_3 & 0.00025595956652173343544298678153836590354330837727 \\ 
GPT2XLtemp0.5 & 0.00024938873842641275406159517835646965977502986789 \\ 
Qwen2CS1.0\_5 & 0.00024654373586945396842548539062534018739825114608 \\ 
GPT2XLtemp0.1 & 0.00024398286403957121048872991320166647710721008480 \\ 
GPT2XLCS0.8\_50 & 0.00024163505809402709297549483036249284850782714784 \\ 
Deepseektemp0.1 & 0.00023924382328677993572382809439602624479448422790 \\ 
Falcon2temp0.9 & 0.00023769853663191438798533983156602289454895071685 \\ 
GPT2XLCS1.0\_50 & 0.00023349758611007252143837442659446423931512981653 \\ 
DeepseekCS1.0\_50 & 0.00023185298386775566020084893281705262779723852873 \\ 
Qwen2CS1.0\_50 & 0.00022417294986173670438833749329887723433785140514 \\ 
Falcon2CS1.0\_1 & 0.00022257922025785455556572134394599515871959738433 \\ 
Qwen2CS1.0\_10 & 0.00022250654320325725064877442349597913562320172787 \\ 
DeepseekCS1.0\_1 & 0.00022207609798458236819058975175522618883405812085 \\ 
Mistral3CS1.0\_50 & 0.00021245058608856940058663698156493637725361622870 \\ 
Deepseektopk1 & 0.00020027873798188301009311207412366684366133995354 \\ 
Qwen2CS1.0\_20 & 0.00019864082380108201994831085279713533964240923524 \\ 
Falcon2CS1.0\_50 & 0.00019670549578724416651086259744118933667778037488 \\ 
GPT2XLtopk10 & 0.00018785668468830190322785278489448046457255259156 \\ 
Deepseektemp0.9 & 0.00016211475163712085701891996158963138441322371364 \\ 
GPT2XLCS1.0\_15 & 0.00014901840847557110505440525205500534866587258875 \\ 
GPT2XLtopk15 & 0.00013414941154746070771548727140753953790408559144 \\ 
GPT2XLCS1.0\_10 & 0.00012462607739036307523405877617506121168844401836 \\ 
GPT2XLtopk20 & 0.00012070069287205829108161614371397263312246650457 \\ 
GPT2XLtopp0.8 & 0.00011867652430858867862364242062866992455383297056 \\ 
GPT2XLtopp0.6 & 0.00011136041634687898814557588389462239319982472807 \\ 
GPT2XLCS1.0\_5 & 0.00009767065751286173476940211113372924955911003053 \\ 
GPT2XLCS1.0\_20 & 0.00008180162601087915977358250296802566481346730143 \\ 

  \hline
\end{tabular}
}
\caption{Estimated worth parameter of the extended Bradley-Terry model based on WikiText-103, Wikinews, and Book datasets together and the metric coherence, diversity, and perplexity (2/2). \label{tab:bt_alldatasets2}}
\end{table}

\section{Discussion of the Ufg-depth Results}

At first glance, this result seems to contradict the number of observations of the partial orders, since the most frequent order, 646 out of 1314, has the lowest depth, and the one with the highest depth is observed only once. But let us take a closer look at the definition of the ufg-depth. The ufg-depth considers subsets of observed partial orders $S$ with size greater than 2, where, in a first step, the number of occurrences is ignored (i.e. not every subset of partial orders is considered, for details see~\cite{blocher2024comparing}). Then, in a second step, the ufg-depth of a partial order is the proportion of the set $S$ that supports that partial order (e.g. the partial order lies between the intersection and union of $S$). This proportion is weighted by the proportion of the number of observations corresponding to the partial orders in~$S$. %
For this dataset, we have that almost all subsets of partial orders do not agree on any dominance structure. Thus, the empty partial order is supported by almost all subsets and, therefore, has such a high depth. Summing things up, the reasons for the low depth value of the most frequent observation are 1) that the number of observations is only considered as a weight and not directly, and 2) that the only subsets $S$ that support this partial order are those that contain the partial order itself in $S$.  Since the partial order corresponding to the highest ufg-depth does not have much in common with other observed partial orders, this set $S$ always implies many other also observed partial orders.\footnote{Note that this observation can also be made for the second (280 out of 1314) and third (208 out of 1314) most observed partial orders .}

\section{Results of Q*Text}\label{a: qtext}

Based on the Q*Text metric introduced in \S\ref{sec:collapsing_single_metric}, we can induce a total ordering of decoding methods. Tables \ref{tab: qtext_md_model}, \ref{tab: qtext_md_strat}, \ref{tab: qtext_md_hyper} and \ref{tab: qtext_md_method} illustrate the results for the most dominant decoding models, strategies, hyperparameters and methods, respectively. On the other hand, We observe in Tables \ref{tab: qtext_ld_model}, \ref{tab: qtext_ld_strat}, \ref{tab: qtext_ld_hyper} and \ref{tab: qtext_ld_method} the results for the least dominant decoding models, strategies, hyperparameters and methods.

\paragraph{Alignment with extended Bradley-Terry}
In this section, we explore the alignment between the extended Bradley-Terry model and Q*Text through various decoding methods.

\begin{figure*}
\centering 
\includegraphics[width=150mm, keepaspectratio]{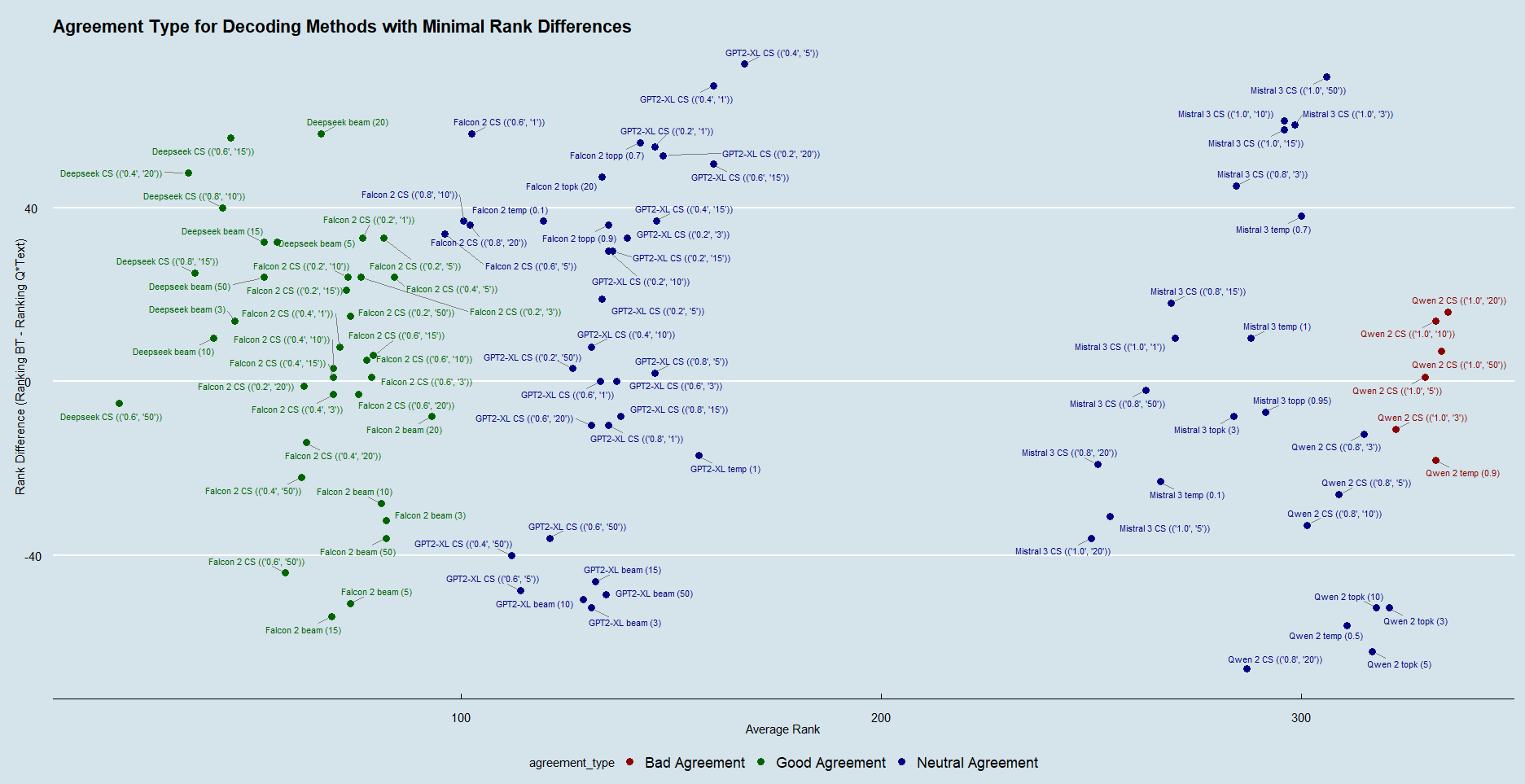} 
\captionof{figure}{Decoding methods with the smallest rank discrepancies between the extended Bradley-Terry model and Q*Text. Green instances represent decoding methods where both rankings agree on high performance; blue instances indicate agreement on neutrality; and red instances signify agreement on lower quality.} 
\label{fig: agreement_bt_qtext} 
\end{figure*}
\begin{figure*}
\centering 
\includegraphics[width=120mm, keepaspectratio]{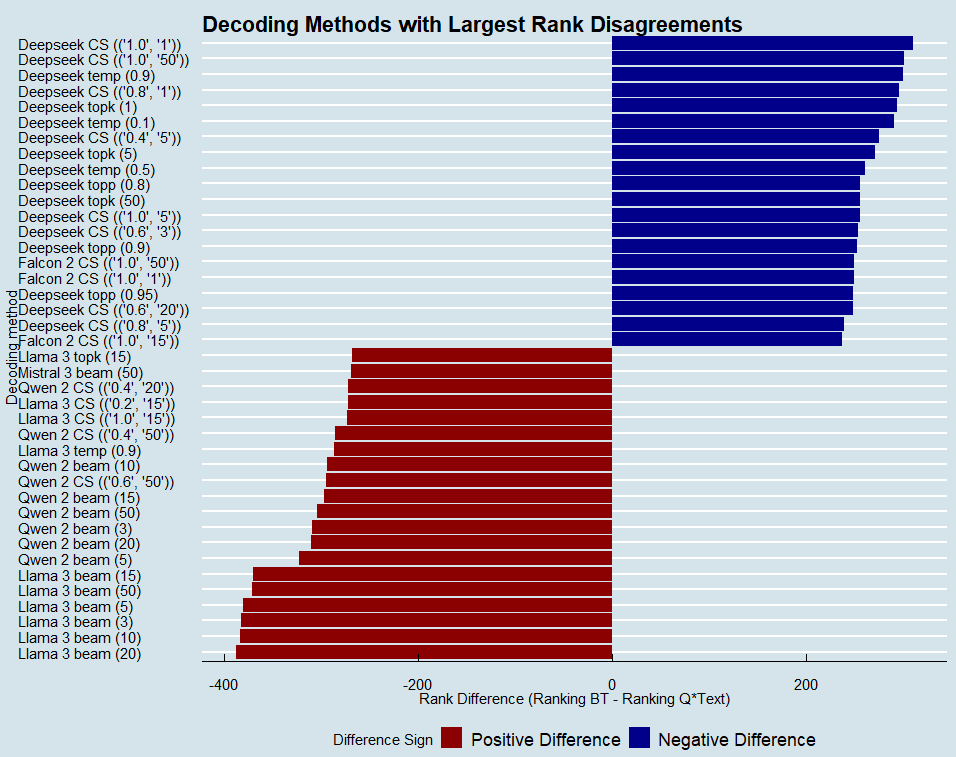} 
\captionof{figure}{Decoding methods with the largest rank discrepancies between the extended Bradley-Terry model and Q*Text. Here, the extended Bradley-Terry model notably favors low-diversity methods, such as BS, while Q*Text tends to rank highly diverse methods higher. This highlights the differing emphases of each approach on diversity in decoding strategies.}
\label{fig: disagreement_bt_qtext} 
\end{figure*}

\clearpage

\begin{table}[H]
    \centering
    \begin{tabular}{lll}
        Most Dominant Model & Count & Proportion \\ \hline
        Falcon 2 & 2195 & 42\% \\ 
        Mistral 3 & 1471 & 28\% \\ 
        Qwen 2 & 904 & 17\% \\ 
        Deepseek & 617 & 12\% \\ 
        GPT2-XL & 55 & 1\% \\ 
        LLama 3 & 19 & 0\% \\ \hline
        Total & 5261 & 100\% \\ 
    \end{tabular}
    \caption{Most dominant models based on Q*Text results.}
        \label{tab: qtext_md_model}
\end{table}

\begin{table}[H]
    \centering
    \begin{tabular}{lll}
        Least Dominant Model & Count & Proportion \\ \hline
        GPT2-XL & 4050 & 77\% \\ 
        Qwen 2 & 703 & 13\% \\ 
        Llama 3 & 259 & 5\% \\ 
        Mistral 3 & 106 & 2\% \\ 
        Deepseek & 80 & 2\% \\ 
        Falcon 2 & 63 & 1\% \\ \hline
        Total & 5261 & 100\% \\ 
    \end{tabular}
    \caption{Least dominant models based on Q*Text results.}
    \label{tab: qtext_ld_model}
\end{table}

\begin{table}[H]
    \centering
    \begin{tabular}{lll}
        Most Dominant Strategy & Count & Proportion \\ \hline
        CS & 5095 & 97\% \\ 
        temp & 135 & 3\% \\ 
        topp & 16 & 0\% \\ 
        topk & 12 & 0\% \\ 
        beam & 3 & 0\% \\ \hline
        Total & 5261 & 100\% \\ 
    \end{tabular}
    \caption{Most dominant strategies based on Q*Text results.}
            \label{tab: qtext_md_strat}
\end{table}

\begin{table}[H]
    \centering
    \begin{tabular}{lll}
        Least Dominant Strategy & Count & Proportion \\ \hline
        CS & 4567 & 87\% \\ 
        beam & 652 & 12\% \\ 
        temp & 34 & 1\% \\ 
        topk & 5 & 0\% \\ 
        topp & 3 & 0\% \\ \hline
        Total & 5261 & 100\% \\ 
    \end{tabular}
    \caption{Least dominant strategies based on Q*Text results.}
        \label{tab: qtext_ld_strat}
\end{table}

\begin{table}[H]
    \centering
    \tiny{
    \begin{tabular}{lll}
        Most Dominant Hyperparameter & Count & Proportion \\ \hline
        ('0.8', '1') & 2138 & 41\% \\ 
        ('1.0', '1') & 830 & 16\% \\ 
        ('0.6', '1') & 805 & 15\% \\ 
        ('0.8', '5') & 360 & 7\% \\ 
        ('0.8', '10') & 216 & 4\% \\ 
        ('0.6', '10') & 163 & 3\% \\ 
        ('0.8', '3') & 89 & 2\% \\ 
        ('0.4', '3') & 86 & 2\% \\ 
        ('0.6', '5') & 71 & 1\% \\ 
        0.7 & 70 & 1\% \\ 
        ('0.8', '15') & 64 & 1\% \\ 
        ('0.6', '3') & 60 & 1\% \\ 
        ('0.4', '10') & 55 & 1\% \\ 
        0.1 & 39 & 1\% \\ 
        ('0.2', '10') & 34 & 1\% \\ 
        ('0.2', '3') & 26 & 0\% \\ 
        0.3 & 22 & 0\% \\ 
        ('0.8', '20') & 18 & 0\% \\ 
        ('0.4', '1') & 17 & 0\% \\ 
        ('0.4', '5') & 13 & 0\% \\ 
        ('0.6', '20') & 12 & 0\% \\ 
        ('0.6', '15') & 11 & 0\% \\ 
        ('1.0', '3') & 8 & 0\% \\ 
        0.5 & 6 & 0\% \\ 
        3 & 6 & 0\% \\ 
        0.9 & 6 & 0\% \\ 
        0.8 & 6 & 0\% \\ 
        ('0.6', '50') & 5 & 0\% \\ 
        10 & 5 & 0\% \\ 
        ('0.2', '1') & 4 & 0\% \\ 
        ('0.2', '20') & 2 & 0\% \\ 
        ('0.2', '5') & 2 & 0\% \\ 
        ('0.4', '20') & 2 & 0\% \\ 
        20 & 2 & 0\% \\ 
        0.6 & 2 & 0\% \\ 
        ('0.4', '15') & 2 & 0\% \\ 
        ('1.0', '5') & 1 & 0\% \\ 
        50 & 1 & 0\% \\ 
        ('0.2', '15') & 1 & 0\% \\ 
        15 & 1 & 0\% \\ \hline
        Total & 5261 & 100\% \\ 
    \end{tabular}
    }
    \caption{Most dominant hyperparameters based on Q*Text results.}
            \label{tab: qtext_md_hyper}    
\end{table}

\begin{table}[H]
    \centering
    \tiny{
    \begin{tabular}{lll}
        Least Dominant Hyperparameter & Count & Proportion \\ \hline
        ('1.0', '50') & 4439 & 0.84 \\ 
        50 & 366 & 0.07 \\ 
        10 & 99 & 0.02 \\ 
        15 & 64 & 0.01 \\ 
        20 & 62 & 0.01 \\ 
        5 & 40 & 0.01 \\ 
        ('1.0', '20') & 39 & 0.01 \\ 
        ('1.0', '15') & 30 & 0.01 \\ 
        ('0.8', '50') & 27 & 0.01 \\ 
        3 & 22 & 0 \\ 
        0.1 & 20 & 0 \\ 
        ('0.2', '1') & 14 & 0 \\ 
        0.3 & 9 & 0 \\ 
        0.5 & 5 & 0 \\ 
        ('0.4', '15') & 5 & 0 \\ 
        1 & 4 & 0 \\ 
        ('0.6', '1') & 3 & 0 \\ 
        ('0.4', '50') & 3 & 0 \\ 
        0.7 & 1 & 0 \\ 
        0.6 & 1 & 0 \\ 
        ('0.2', '10') & 1 & 0 \\ 
        0.95 & 1 & 0 \\ 
        ('0.6', '5') & 1 & 0 \\ 
        ('0.8', '10') & 1 & 0 \\ 
        ('0.6', '20') & 1 & 0 \\ 
        ('0.4', '5') & 1 & 0 \\ 
        ('0.4', '3') & 1 & 0 \\ 
        ('0.2', '15') & 1 & 0 \\ \hline
        Total & 5261 & 100\% \\ 
    \end{tabular}
    }
    \caption{Least dominant hyperparameters based on Q*Text results.}
            \label{tab: qtext_ld_hyper}
\end{table}

\begin{table}[H]
    \centering
        {\tiny
        \begin{tabular}{lll}
        Most Dominant Method & Count & Proportion \\ \hline
        Falcon 2\_CS (('0.8', '1')) & 1083 & 21\% \\ 
        Mistral 3\_CS (('0.8', '1')) & 656 & 12\% \\ 
        Mistral 3\_CS (('0.6', '1')) & 629 & 12\% \\ 
        Falcon 2\_CS (('1.0', '1')) & 510 & 10\% \\ 
        Falcon 2\_CS (('0.8', '5')) & 335 & 6\% \\ 
        Qwen 2\_CS (('0.8', '1')) & 317 & 6\% \\ 
        Deepseek\_CS (('0.6', '1')) & 160 & 3\% \\ 
        Qwen 2\_CS (('0.8', '10')) & 148 & 3\% \\ 
        Deepseek\_CS (('1.0', '1')) & 141 & 3\% \\ 
        Qwen 2\_CS (('1.0', '1')) & 112 & 2\% \\ 
        Falcon 2\_CS (('0.6', '10')) & 99 & 2\% \\ 
        Deepseek\_CS (('0.8', '1')) & 76 & 1\% \\ 
        Deepseek\_CS (('0.4', '3')) & 70 & 1\% \\ 
        Falcon 2\_CS (('0.8', '10')) & 68 & 1\% \\ 
        Falcon 2\_CS (('0.6', '5')) & 67 & 1\% \\ 
        Qwen 2\_CS (('0.8', '15')) & 63 & 1\% \\ 
        Deepseek\_CS (('0.6', '10')) & 58 & 1\% \\ 
        Qwen 2\_CS (('0.4', '10')) & 48 & 1\% \\ 
        Mistral 3\_temp (0.7) & 45 & 1\% \\ 
        GPT2-XL\_CS (('1.0', '1')) & 42 & 1\% \\ 
        Qwen 2\_CS (('0.8', '3')) & 41 & 1\% \\ 
        Deepseek\_CS (('0.8', '3')) & 37 & 1\% \\ 
        Qwen 2\_CS (('0.2', '10')) & 32 & 1\% \\ 
        Mistral 3\_CS (('0.6', '3')) & 31 & 1\% \\ 
        Mistral 3\_CS (('1.0', '1')) & 30 & 1\% \\ 
        Mistral 3\_temp (0.1) & 29 & 1\% \\ 
        Qwen 2\_CS (('0.6', '3')) & 20 & 0\% \\ 
        Falcon 2\_CS (('0.6', '1')) & 19 & 0\% \\ 
        Deepseek\_CS (('0.2', '3')) & 19 & 0\% \\ 
        Deepseek\_CS (('0.4', '1')) & 17 & 0\% \\ 
        Qwen 2\_CS (('0.8', '20')) & 15 & 0\% \\ 
        Qwen 2\_CS (('0.8', '5')) & 15 & 0\% \\ 
        Qwen 2\_CS (('0.4', '3')) & 15 & 0\% \\ 
        Mistral 3\_CS (('0.8', '3')) & 14 & 0\% \\ 
        Qwen 2\_CS (('0.6', '15')) & 12 & 0\% \\ 
        Mistral 3\_temp (0.3) & 12 & 0\% \\ 
        Mistral 3\_CS (('0.4', '5')) & 11 & 0\% \\ 
        Deepseek\_CS (('0.6', '3')) & 11 & 0\% \\ 
        GPT2-XL\_CS (('0.8', '1')) & 10 & 0\% \\ 
        Falcon 2\_temp (0.7) & 10 & 0\% \\ 
        Qwen 2\_topp (0.7) & 9 & 0\% \\ 
        Qwen 2\_CS (('0.6', '10')) & 9 & 0\% \\ 
        Qwen 2\_temp (0.7) & 9 & 0\% \\ 
        Mistral 3\_CS (('0.8', '5')) & 8 & 0\% \\ 
        Qwen 2\_temp (0.3) & 7 & 0\% \\ 
        Qwen 2\_CS (('0.2', '3')) & 7 & 0\% \\ 
        Qwen 2\_temp (0.9) & 7 & 0\% \\ 
        Deepseek\_CS (('0.4', '10')) & 7 & 0\% \\ 
        Qwen 2\_temp (0.1) & 7 & 0\% \\ 
        Mistral 3\_CS (('0.6', '20')) & 6 & 0\% \\ 
        Deepseek\_CS (('0.8', '5')) & 6 & 0\% \\ 
        Deepseek\_CS (('0.6', '5')) & 6 & 0\% \\ 
        Qwen 2\_topk (3) & 6 & 0\% \\ 
        Qwen 2\_CS (('1.0', '3')) & 6 & 0\% \\ 
        Deepseek\_temp (0.5) & 5 & 0\% \\ 
        Falcon 2\_CS (('0.8', '20')) & 5 & 0\% \\ 
        Deepseek\_CS (('0.2', '1')) & 5 & 0\% \\ 
        Qwen 2\_topp (0.8) & 5 & 0\% \\ 
        Qwen 2\_topk (10) & 5 & 0\% \\ 
        Deepseek\_temp (0.1) & 5 & 0\% \\ 
        LLama 3\_temp (0.3) & 4 & 0\% \\ \hline
        Total & 5261 & 100\% \\ 
    \end{tabular}
    }
    \caption{Most dominant methods based on Q*Text results.}
    \label{tab: qtext_md_method}
\end{table}

\begin{table}[H]
    \centering
    {\tiny
    \begin{tabular}{lll}
        Least Dominant Method & Count & Proportion \\ \hline
        GPT2-XL\_CS (('1.0', '50')) & 3821 & 73\% \\ 
        Qwen 2\_CS (('1.0', '50')) & 561 & 11\% \\ 
        LLama 3\_beam (50) & 130 & 2\% \\ 
        GPT2-XL\_beam (50) & 95 & 2\% \\ 
        Qwen 2\_beam (50) & 53 & 1\% \\ 
        Mistral 3\_beam (50) & 51 & 1\% \\ 
        LLama 3\_beam (10) & 38 & 1\% \\ 
        GPT2-XL\_beam (10) & 38 & 1\% \\ 
        Deepseek\_CS (('1.0', '50')) & 34 & 1\% \\ 
        Qwen 2\_CS (('1.0', '20')) & 29 & 1\% \\ 
        GPT2-XL\_CS (('1.0', '15')) & 29 & 1\% \\ 
        LLama 3\_beam (20) & 27 & 1\% \\ 
        LLama 3\_beam (15) & 26 & 0\% \\ 
        Deepseek\_beam (50) & 22 & 0\% \\ 
        LLama 3\_beam (5) & 18 & 0\% \\ 
        Mistral 3\_CS (('1.0', '50')) & 16 & 0\% \\ 
        Qwen 2\_beam (10) & 15 & 0\% \\ 
        Falcon 2\_beam (50) & 15 & 0\% \\ 
        Qwen 2\_CS (('0.8', '50')) & 15 & 0\% \\ 
        Mistral 3\_beam (15) & 14 & 0\% \\ 
        GPT2-XL\_beam (20) & 10 & 0\% \\ 
        GPT2-XL\_beam (5) & 10 & 0\% \\ 
        GPT2-XL\_beam (3) & 9 & 0\% \\ 
        Falcon 2\_CS (('1.0', '20')) & 9 & 0\% \\ 
        GPT2-XL\_CS (('0.2', '1')) & 8 & 0\% \\ 
        Qwen 2\_beam (15) & 8 & 0\% \\ 
        Mistral 3\_beam (20) & 8 & 0\% \\ 
        Qwen 2\_beam (20) & 7 & 0\% \\ 
        Deepseek\_beam (20) & 7 & 0\% \\ 
        Falcon 2\_CS (('1.0', '50')) & 7 & 0\% \\ 
        Falcon 2\_CS (('0.8', '50')) & 7 & 0\% \\ 
        LLama 3\_temp (0.1) & 6 & 0\% \\ 
        Deepseek\_beam (15) & 6 & 0\% \\ 
        GPT2-XL\_beam (15) & 5 & 0\% \\ 
        GPT2-XL\_CS (('0.4', '15')) & 5 & 0\% \\ 
        Falcon 2\_beam (15) & 5 & 0\% \\ 
        Qwen 2\_beam (3) & 5 & 0\% \\ 
        GPT2-XL\_temp (0.1) & 5 & 0\% \\ 
        GPT2-XL\_CS (('0.8', '50')) & 4 & 0\% \\ 
        Mistral 3\_beam (5) & 4 & 0\% \\ 
        Mistral 3\_beam (3) & 4 & 0\% \\ 
        Mistral 3\_temp (0.1) & 3 & 0\% \\ 
        LLama 3\_temp (0.3) & 3 & 0\% \\ 
        GPT2-XL\_temp (0.3) & 3 & 0\% \\ 
        Falcon 2\_temp (0.1) & 3 & 0\% \\ 
        Mistral 3\_beam (10) & 3 & 0\% \\ 
        Falcon 2\_beam (20) & 3 & 0\% \\ 
        GPT2-XL\_CS (('0.6', '1')) & 3 & 0\% \\ 
        Deepseek\_beam (10) & 3 & 0\% \\ 
        Falcon 2\_beam (5) & 3 & 0\% \\ 
        Qwen 2\_beam (5) & 3 & 0\% \\ 
        Mistral 3\_temp (0.3) & 2 & 0\% \\ 
        Qwen 2\_temp (0.1) & 2 & 0\% \\ 
        Falcon 2\_beam (10) & 2 & 0\% \\ 
        Deepseek\_topk (1) & 2 & 0\% \\ 
        LLama 3\_beam (3) & 2 & 0\% \\ 
        LLama 3\_CS (('0.2', '1')) & 2 & 0\% \\ 
        Qwen 2\_CS (('0.2', '1')) & 2 & 0\% \\ 
        Deepseek\_beam (5) & 2 & 0\% \\ 
        Falcon 2\_topk (1) & 2 & 0\% \\ 
        Falcon 2\_temp (0.5) & 2 & 0\% \\ 
        GPT2-XL\_temp (0.5) & 2 & 0\% \\ 
        Qwen 2\_topp (0.7) & 1 & 0\% \\ 
        Qwen 2\_temp (0.3) & 1 & 0\% \\ 
        LLama 3\_CS (('0.6', '20')) & 1 & 0\% \\ 
        LLama 3\_temp (0.5) & 1 & 0\% \\ 
        Deepseek\_temp (0.1) & 1 & 0\% \\ 
        Falcon 2\_CS (('0.4', '5')) & 1 & 0\% \\ 
        GPT2-XL\_CS (('0.2', '10')) & 1 & 0\% \\ 
        GPT2-XL\_topp (0.95) & 1 & 0\% \\ 
        LLama 3\_CS (('0.8', '50')) & 1 & 0\% \\ 
        LLama 3\_CS (('0.6', '5')) & 1 & 0\% \\ 
        LLama 3\_CS (('0.8', '10')) & 1 & 0\% \\ 
        Deepseek\_CS (('0.4', '50')) & 1 & 0\% \\ 
        Qwen 2\_CS (('0.4', '50')) & 1 & 0\% \\ 
        LLama 3\_topp (0.6) & 1 & 0\% \\ 
        GPT2-XL\_topk (3) & 1 & 0\% \\ 
        Falcon 2\_CS (('0.4', '50')) & 1 & 0\% \\ 
        Falcon 2\_CS (('0.4', '3')) & 1 & 0\% \\ 
        Deepseek\_CS (('1.0', '15')) & 1 & 0\% \\ 
        LLama 3\_CS (('0.2', '15')) & 1 & 0\% \\ 
        Falcon 2\_CS (('0.2', '1')) & 1 & 0\% \\ 
        Falcon 2\_beam (3) & 1 & 0\% \\ 
        Deepseek\_CS (('1.0', '20')) & 1 & 0\% \\ 
        Mistral 3\_CS (('0.2', '1')) & 1 & 0\% \\ \hline
        Total & 5261 & 100\% \\ 
    \end{tabular}
    }
    \caption{Least dominant methods based on Q*Text results.}
                \label{tab: qtext_ld_method}
\end{table}

\clearpage

\section{Q*Text Hyperparameters}
\label{a:qtext_parameters}

\begin{table}[htbp]
\centering
\resizebox{\columnwidth}{!}{
\begin{tabular}{rl}
\toprule
\textbf{Line} & \textbf{Pseudocode: Q*Text Hyperparameter Tuning} \\
& \textbf{Input:} Perplexity, Coherence and Diversity scores (P, C, D)\\
\midrule
1 & \texttt{P\_norm = (max(P) - P) / (max(P) - min(P))} \\
2 & \texttt{C\_norm = (C - min(C)) / (max(C) - min(C))} \\
3 & \texttt{D\_norm = (D - min(D)) / (max(D) - min(D))} \\
4 & \texttt{$\theta$ = [1,1,1,0.5,0.5,0.5,1,1,1]} \\
5 & \texttt{bounds\_w = [[0.1,5],[0.1,5],[0.1,5]]} \\
6 & \texttt{bounds\_$\mu$ = [[0,1],[0,1],[0,1]]} \\
7 & \texttt{bounds\_$\alpha$ = [[0.1,10],[0.1,10],[0.1,10]]} \\
8 & \texttt{for trial in range(max\_trials):} \\
9 & \quad \texttt{$\theta$\_new = $\theta$ + random\_normal(0, 0.1)} \\
10 & \quad \texttt{$\theta$\_new = clip($\theta$\_new, bounds)} \\
11 & \quad \texttt{for i in range(N):} \\
12 & \quad\quad \texttt{penalty\_p = exp(-$\alpha_1$(P\_norm[i]-$\mu_1$)$^2$)} \\
13 & \quad\quad \texttt{penalty\_c = exp(-$\alpha_2$(C\_norm[i]-$\mu_2$)$^2$)} \\
14 & \quad\quad \texttt{penalty\_d = exp(-$\alpha_3$(D\_norm[i]-$\mu_3$)$^2$)} \\
15 & \quad\quad \texttt{QText[i] = ($w_1$P\_norm[i]penalty\_p + } \\
16 & \quad\quad\quad\quad\quad\quad \texttt{$w_2$C\_norm[i]penalty\_c + } \\
17 & \quad\quad\quad\quad\quad\quad \texttt{$w_3$D\_norm[i]penalty\_d) / ($w_1$+$w_2$+$w_3$)} \\
18 & \quad \texttt{$\rho$ = spearman\_corr(QText, Human)} \\
19 & \quad \texttt{if $\rho$ > best\_$\rho$: $\theta$\_best = $\theta$\_new} \\
20 & \texttt{return $\theta$\_best} \\
\bottomrule
\end{tabular}%
}
\caption{Q*Text Optimization Algorithm}
\label{tab:qtext_algorithm}
\end{table}

\noindent\textit{Algorithm explanation:} Lines 1-3 normalize metrics to [0,1]. Lines 5-7 define parameter bounds for weights ($w_i \in [0.1, 5.0]$), targets ($\mu_i \in [0.0, 1.0]$), and penalties ($\alpha_i \in [0.1, 10.0]$), this bound definition aims at (i) preventing zero weights while allowing one metric to dominate, (ii) match the normalized metric range, and (iii) ensure positive penalties with reasonable strength. Lines 9-10 perturb parameters with Gaussian noise and clip to bounds. The optimization maximizes Spearman correlation $\rho$ with human ratings.

\begin{table}[htbp]
\centering
\begin{tabular}{lcc}
\toprule
\textbf{Parameter} & \textbf{Symbol} & \textbf{Value} \\
\midrule
\multicolumn{3}{l}{\textit{Metric Weights}} \\
Perplexity Weight & $w_1$ & 0.586 \\
Coherence Weight & $w_2$ & 0.834 \\
Diversity Weight & $w_3$ & 3.853 \\
\midrule
\multicolumn{3}{l}{\textit{Gaussian Target Values ($\mu$)}} \\
Perplexity Target & $\mu_1$ & 0.458 \\
Coherence Target & $\mu_2$ & 0.000 \\
Diversity Target & $\mu_3$ & 0.854 \\
\midrule
\multicolumn{3}{l}{\textit{Gaussian Penalty Strength ($\alpha$)}} \\
Perplexity Penalty & $\alpha_1$ & 2.579 \\
Coherence Penalty & $\alpha_2$ & 1.496 \\
Diversity Penalty & $\alpha_3$ & 7.370 \\
\bottomrule
\end{tabular}
\caption{Optimal Q*Text Hyperparameters (Spearman $\rho_s = 0.5545$)}
\label{tab:qtext_parameters}
\end{table}

\paragraph{Parameter Interpretation.} The optimized parameters reveal insights about text quality assessment.\\
\noindent\textbf{Diversity dominance:} The substantially higher weight for diversity ($w_3 = 3.853$) compared to perplexity ($w_1 = 0.586$) and coherence ($w_2 = 0.834$) indicates that lexical variety is the most discriminative factor for human preferences in our dataset.\\
\noindent\textbf{Target preferences:} The optimal targets suggest humans prefer moderate perplexity levels ($\mu_1 = 0.458$), minimal coherence constraints ($\mu_2 = 0.000$), and high diversity ($\mu_3 = 0.854$).\\
\noindent\textbf{Penalty sensitivity:} The high diversity penalty strength ($\alpha_3 = 7.370$) enforces strict adherence to the diversity target, while the moderate perplexity penalty ($\alpha_1 = 2.579$) and lenient coherence penalty ($\alpha_2 = 1.496$) allow more variation in these two dimensions.

\begin{figure}[htbp]
\centering
\includegraphics[width=1\columnwidth]{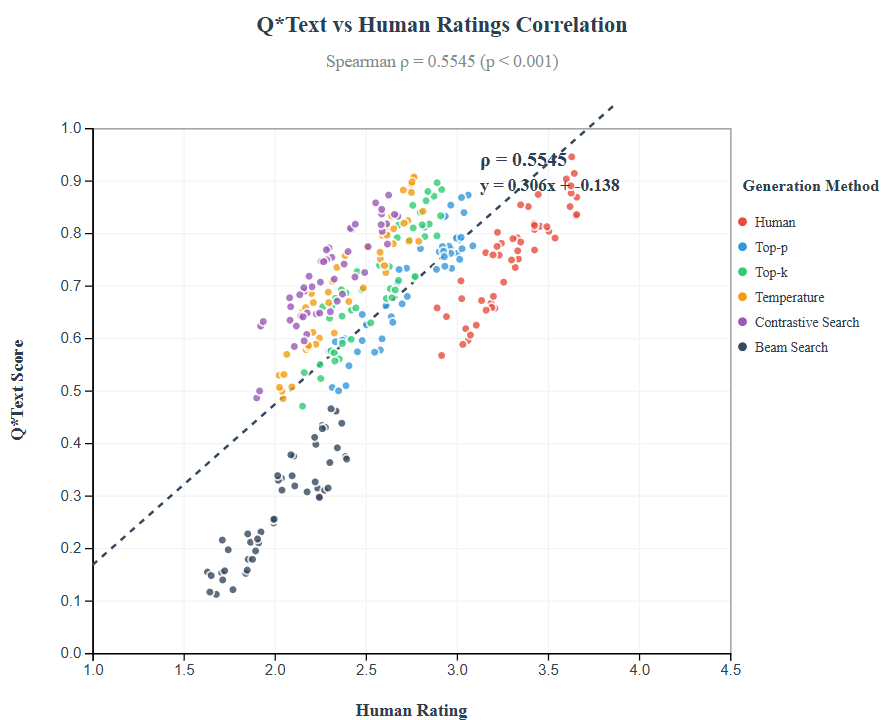}
\caption{Correlation between Q*Text scores and human ratings across six text generation methods. Each point represents a text sample, colored by generation method. The dashed line shows the linear regression fit. Q*Text achieves a moderate positive correlation (Spearman $\rho = 0.5545$, $p < 0.001$) with human evaluations, demonstrating its effectiveness in capturing human preferences for text quality.}
\label{fig:qtext_correlation}
\end{figure}




\end{document}